\pdfoutput=1

\documentclass[11pt]{article}

\usepackage{acl}    

\usepackage{times}
\usepackage{latexsym}

\usepackage[T1]{fontenc}

\usepackage[utf8]{inputenc}

\usepackage{microtype}

\usepackage{inconsolata}

\usepackage{booktabs}       
\usepackage{amsfonts}       
\usepackage{nicefrac}       
\usepackage{xcolor}         
\usepackage{wrapfig}
\usepackage{nicematrix}
\usepackage{mathtools}
\usepackage{multirow}
\usepackage{makecell}
\usepackage{array}
\usepackage{subcaption}
\usepackage{tabularx,colortbl}
\usepackage{enumitem}
\usepackage{commath}
\usepackage{soul}
\usepackage{amsmath}
\usepackage{float}
\usepackage{listings}
\usepackage{tcolorbox}
\usepackage{xspace}
\usepackage{cleveref}
\usepackage{multirow}
\usepackage{newfloat}
\usepackage{tablefootnote}
\usepackage[flushleft]{threeparttable}
\usepackage{seqsplit}
\usepackage{placeins}
\usepackage{times}
\usepackage{latexsym}
\usepackage{amssymb, algorithm, algpseudocode}
\usepackage{MnSymbol} 

\usepackage{xcolor}
\usepackage{pifont}

\usepackage{hyperref}
\usepackage{url}
\usepackage{amsmath,amsfonts,bm}

\def\eqref#1{equation~\ref{#1}}

\def\1{\bm{1}}

\def\vp{{\bm{p}}}
\def\vq{{\bm{q}}}

\def\vs{{\bm{s}}}
\def\vt{{\bm{t}}}

\DeclareMathAlphabet{\mathsfit}{\encodingdefault}{\sfdefault}{m}{sl}
\SetMathAlphabet{\mathsfit}{bold}{\encodingdefault}{\sfdefault}{bx}{n}

\newcommand{\softmax}{\mathrm{softmax}}

\DeclareMathOperator*{\argmax}{arg\,max}

\newcommand{\subsec}[1]{\noindent\textbf{#1}~~}
\definecolor{ForestGreen}{RGB}{34, 139, 34}
\definecolor{TealBlue}{RGB}{0, 204, 204}

\newcommand{\increasenoparent}[1]{\textcolor{ForestGreen}{+#1}}

\newcommand{\decreasenoparent}[1]{\textcolor{red}{-#1}}

\newcommand{\greencheck}{\textcolor{ForestGreen}{\ding{51}}}
\newcommand{\redcross}{\textcolor{red!70!black}{\ding{55}}}

\newcommand{\redcheck}{\textcolor{red!70!black}{\ding{51}}}
\newcommand{\greencross}{\textcolor{ForestGreen}{\ding{55}}}

\newcommand{\extraresults}[1]{(\textcolor{magenta}{#1})}

\crefname{section}{Sec.}{Secs.}
\crefname{table}{Table}{Tables}
\crefname{figure}{Fig.}{Figs.}

\definecolor{myGray}{rgb}{0.94,0.94,0.94}
\definecolor{magenta(dye)}{rgb}{0.79, 0.08, 0.48}
\definecolor{mycitecolor}{rgb}{0,0.08,0.75} %
\definecolor{arrow_blue}{rgb}{0,0.44,0.75} 
\definecolor{codegray}{rgb}{0.5,0.5,0.5}
\definecolor{codepurple}{rgb}{0.58,0,0.82}
\definecolor{backcolour}{rgb}{0.95,0.95,0.92}

\definecolor{darkmagenta}{RGB}{165, 0, 64}
\definecolor{darkcyan}{RGB}{0, 110, 175}

\newcommand{\dog}{\includegraphics[scale=0.11]{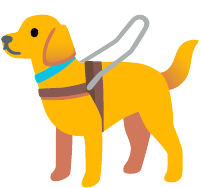}\xspace}
\newcommand{\bird}{\includegraphics[scale=0.11]{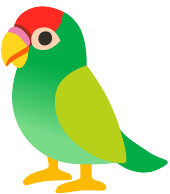}\xspace}

\newcommand{\hlgray}[1]{\sethlcolor{lightgray}\hl{#1}}

\newcommand{\eg}{e.g.\xspace}
\newcommand{\ie}{i.e.\xspace}
\newcommand{\class}[1]{{\texttt{#1}}}
\newcommand{\peeb}{PEEB\xspace}
\newcommand{\mv}{M\&V\xspace}
\newcommand{\chatgpt}{GPT-4\xspace}
\newcommand{\birdsoup}{Bird-11K\xspace}
\newcommand{\dogsoup}{Dog-140\xspace}
\newcommand{\owlvit}{OWL-ViT\xspace}
\newcommand{\owlvitBase}{OWL-ViT$_{\text{\ensuremath{\mathsf{B/32}}}}$\xspace}
\newcommand{\owlvitBaseSixteen}{OWL-ViT$_{\text{\ensuremath{\mathsf{B/16}}}}$\xspace}
\newcommand{\owlvitLarge}{OWL-ViT$_{\text{\ensuremath{\mathsf{Large}}}}$\xspace}
\newcommand{\gzsl}{GZSL\xspace}
\newcommand{\zsl}{ZSL\xspace}

\newcommand{\birdsoupLvOne}{Bird-11K$_{\text{\ensuremath{\mathsf{[-test]}}}}$\xspace}
\newcommand{\birdsoupLvThreeCub}{Bird-11K$_{\text{\ensuremath{\mathsf{[-CUB]}}}}$\xspace}
\newcommand{\birdsoupLvThreeNABirds}{Bird-11K$_{\text{\ensuremath{\mathsf{[-NAB]}}}}$\xspace}

\newcommand{\xclipLvOne}{{PEEB}$_{\text{\ensuremath{\mathsf{[-test]}}}}$\xspace}
\newcommand{\xclipLvThreeCUB}{{PEEB}$_{\text{\ensuremath{\mathsf{[-CUB]}}}}$\xspace}
\newcommand{\xclipLvThreeNABirds}{PEEB$_{\text{\ensuremath{\mathsf{[-NAB]}}}}$\xspace}

\newcommand{\xclipLvOneCUB}{PEEB$_{\text{\ensuremath{\mathsf{[-test]}}}}^{\text{\ensuremath{\mathsf{CUB}}}}$\xspace}
\newcommand{\xclipLvThreeCUBClore}{{PEEB}$_{\text{\ensuremath{\mathsf{[-cub]}}}}^{\text{\ensuremath{\mathsf{Akata}}}}$\xspace}
\newcommand{\xclipLvThreeCUBSCS}{{PEEB}$_{\text{\ensuremath{\mathsf{[-cub]}}}}^{\text{\ensuremath{\mathsf{SCS}}}}$\xspace}
\newcommand{\xclipLvThreeCUBSCE}{{PEEB}$_{\text{\ensuremath{\mathsf{[-cub]}}}}^{\text{\ensuremath{\mathsf{SCE}}}}$\xspace}
\newcommand{\xclipLvThreeNABirdsSCS}{PEEB$_{\text{\ensuremath{\mathsf{[-nab]}}}}^{\text{\ensuremath{\mathsf{SCS}}}}$\xspace}
\newcommand{\xclipLvThreeNABirdsSCE}{PEEB$_{\text{\ensuremath{\mathsf{[-nab]}}}}^{\text{\ensuremath{\mathsf{SCE}}}}$\xspace}

\newcommand{\na}{\textcolor{lightgray}{n/a}\xspace}
\newcommand{\cub}{CUB\xspace}
\newcommand{\nabirds}{NABirds\xspace}
\newcommand{\inat}{iNaturalist\xspace}

\newcommand{\cloreClip}{CLORE$_{CLIP}$\xspace}
\newcommand{\ssgadet}{S$^2$GA-DET\xspace}
\newcommand{\grzsl}{GRZSL\xspace}
\newcommand{\dgrzsl}{DGRZSL\xspace}
\newif\ifcomments

\commentstrue

\ifcomments
\newcommand{\comments}[1]{#1}
\else
\newcommand{\comments}[1]{}
\fi

\soulregister{\ref}{7}
\soulregister{\cref}{7}
\soulregister{\Cref}{7}
\soulregister{\citep}{7}
\soulregister{\owlvitBase}{7}
\soulregister{\birdsoup}{7}
\soulregister{\cub}{7}
\soulregister{\nabirds}{7}
\soulregister{\subsec}{7}
\soulregister{\subsection}{7}
\soulregister{\peeb}{7}
\soulregister{\textbf}{7}
\soulregister{\ie}{7}

\newenvironment{prompt}
  {\list{}{\leftmargin=1em \rightmargin=1em \small\ttfamily} \item\relax}
  {\endlist}

\lstdefinestyle{nohighlight}{
    basicstyle=\footnotesize\ttfamily, 
    breaklines=true,                    
    breakatwhitespace=true,            
    tabsize=2,                          
    xleftmargin=0pt,                    
    xrightmargin=0pt,                   
    frame=None                        
}

\makeatletter
\newcommand{\printfnsymbol}[1]{%
  \textsuperscript{\@fnsymbol{#1}}%
}
\makeatother

\newcommand{\papertitle}{PEEB: Part-based Image Classifiers with an \\ Explainable and Editable Language Bottleneck}
\title{\papertitle}

\author{
    Thang M. Pham\thanks{Equal contribution}$^\dagger$ \\
    \texttt{thangpham@auburn.edu} \\
    \And
    Peijie Chen\printfnsymbol{1}$^\dagger$\\
    \texttt{pzc0018@auburn.edu} \\
    \And
    Tin Nguyen\printfnsymbol{1}$^\dagger$\\
    \texttt{ttn0011@auburn.edu} \\
    \AND
    Seunghyun Yoon $^\mathsection$ \\
    \texttt{syoon@adobe.com} \\
    \And
    Trung Bui $^\mathsection$\\
    \texttt{bui@adobe.com} \\
    \And
    Anh Totti Nguyen $^\dagger$\\
   \texttt{anh.ng8@gmail.com}
   \AND
    $^\dagger$\textnormal{Auburn University} ~~~~~ $^\mathsection$\textnormal{Adobe Research}
}

\begin{document}
\maketitle
\begin{abstract}

CLIP-based classifiers rely on the prompt containing a \texttt{\{class name\}} that is known to the text encoder.
Therefore, they perform poorly with new classes or the classes whose names rarely appear on the Internet (\eg, scientific names of birds).
For fine-grained classification, we propose \peeb---an explainable and editable classifier to (1) express the class name into a set of text descriptors that describe the visual parts of the class; and (2) match the embeddings of the detected parts with their textual descriptors in each class to compute a logit score for classification.
In a zero-shot setting where the class names are \emph{unknown},
\peeb significantly outperforms CLIP, achieving a 10-fold increase in top-1 accuracy.
Compared to part-based classifiers, \peeb not only achieves state-of-the-art (SOTA) accuracy in the supervised-learning setting---88.80\% and 92.20\% accuracy on CUB-200 \bird~and Dogs-120 \dog, respectively---but also the first to enable users to \emph{edit} the text descriptors to form a new classifier without any re-training.
Compared to concept bottleneck models, \peeb is also the SOTA in both zero-shot and supervised learning settings.


\end{abstract}

\section{Introduction}
\label{sec:introduction}

\begin{figure}[t]
\centering
\setlength\tabcolsep{1.0pt}



\centering
\includegraphics[width=1\linewidth]{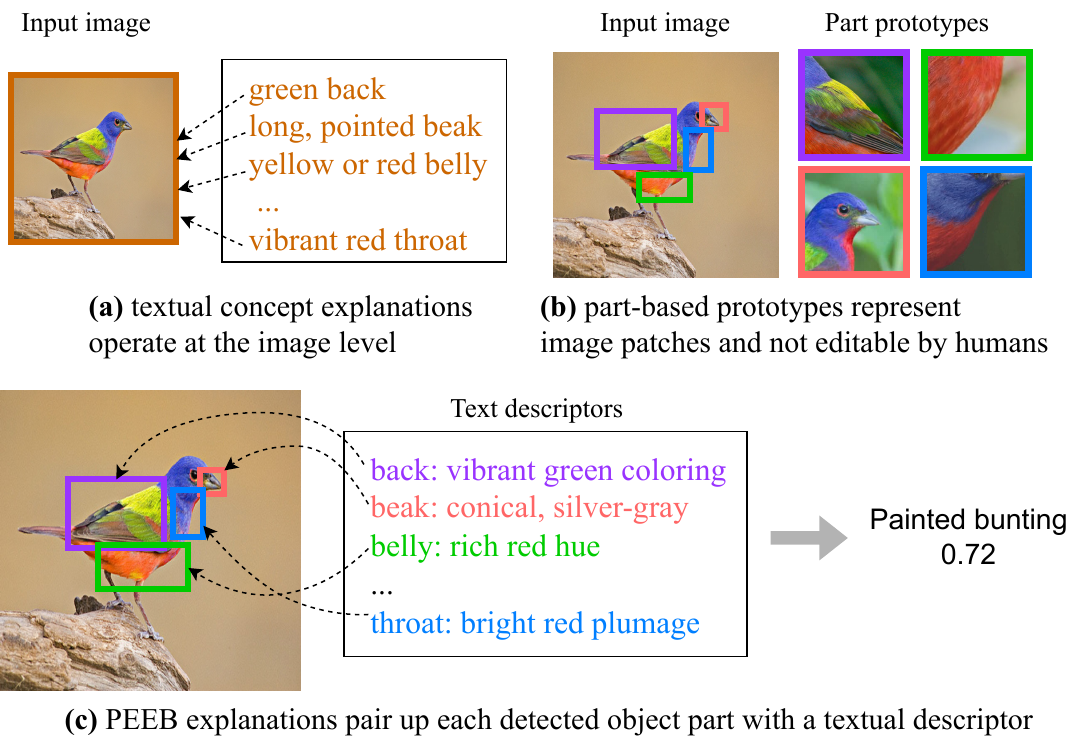}

\caption{
Existing explanations are either (a) textual but at the image level; or (b) part-level but not textual.
Combining the best of both worlds, \peeb (c) first matches each detected object part to a text descriptor, then uses the part-level matching scores to classify the image.
}
\label{fig:teaser}
\end{figure}

Fine-grained classification \citep{wah2011caltech,van2015building} is a long-standing computer-vision challenge.
Furthermore, it is also important to explain how SOTA classifiers make a decision, \eg, which bird traits make a model think a given bird is \class{Painted Bunting}? (\cref{fig:teaser})

First, many bird classifiers claim to be explainable \citep{chen2019looks, donnelly2022deformable} by comparing the input image with a set of learned, part prototypes (\Cref{fig:teaser}b) or natural-language concepts (\Cref{fig:teaser}a).
Yet, such prototypes are feature vectors and therefore not editable by users.
Textual concepts are often compared against \emph{entire image} for classification and it is unknown what image details match a given descriptor \cite{menon2023visual,yang2023language}.
Third, most vision-language classifiers need the prompt to have
a known \texttt{\{class name\}} (like a special code instead of an expressive, natural description) that matches the input image \cite{roth2023waffling}.
Fourth, most classifiers require either training-set images in a supervised-learning setting or demonstration images in a zero-shot setting \citep{xian2018zero,zhu2018generative}.
This requirement is impractical when building a classifier for a novel species whose photos do not yet exist in the database.

To address the above four problems, we propose \peeb, a Part-based image classifier that is Explainable and Editable via a natural-language Bottleneck.
\peeb classifies images by grounding the textual descriptor of object \emph{parts} provided by humans or GPT-4 (no images needed) to detected parts in the image (\cref{fig:teaser}c).
While PEEB leverages CLIP's encoders \citep{radford2021learning}, it uses no class names (\eg, \class{Indigo Bunting}) in the prompt.
In contrast, CLIP-based models \citep{radford2021learning,pratt2022does,menon2023visual} rely so heavily on the \emph{known} class names that their accuracy drops significantly when the names are removed or replaced by less-common ones (\Cref{sec:effect_on_descriptions}).

\begin{figure*}[t]
    \centering
    \includegraphics[width=1.0\linewidth]{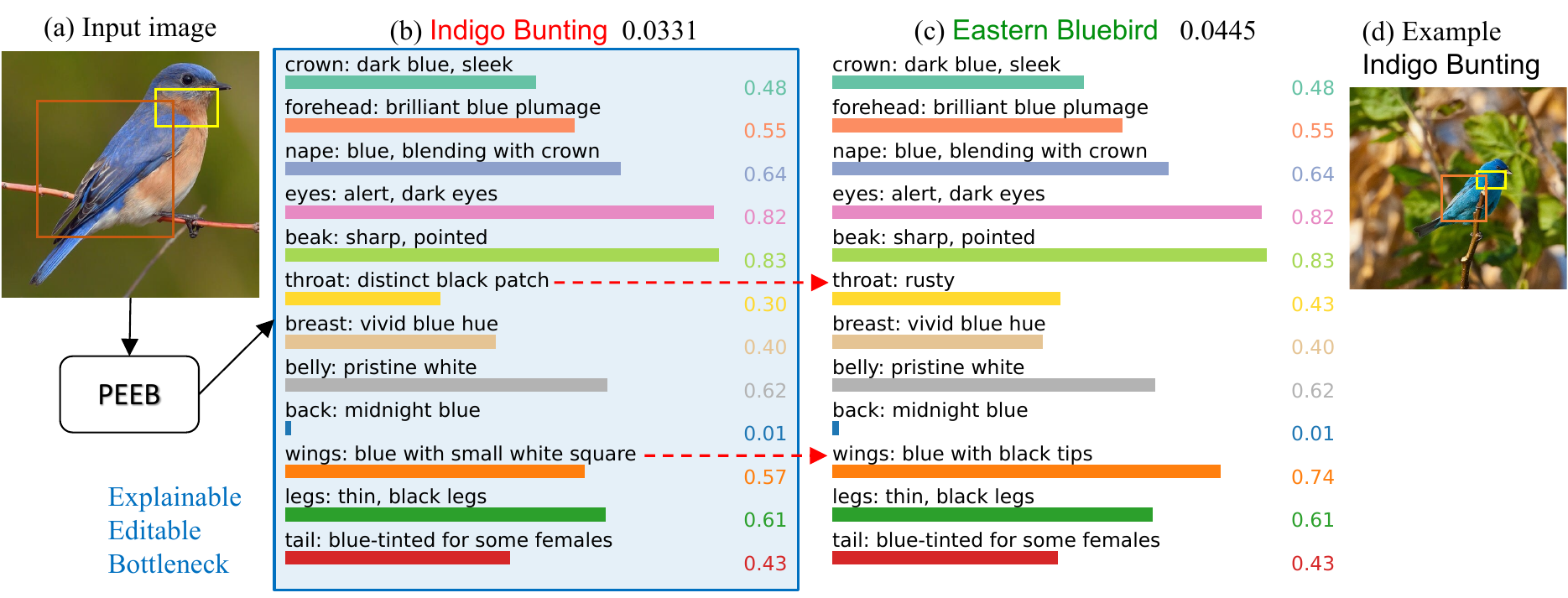}
    \caption{
    Given an input image (a) from an unseen class of \class{Eastern Bluebird}, PEEB misclassifies it into \textcolor{red}{\class{Indigo Bunting}} (b), a visually similar blue bird in CUB-200 (d).
    To add a new class for \class{Eastern Bluebird} to the 200-class list that PEEB considers when classifying, we clone the \textcolor{arrow_blue}{12 textual descriptors} of \class{Indigo Bunting} (b) and \textbf{edit} (\textcolor{red}{- -$\blacktriangleright$}) the {descriptor} of 
    {\sethlcolor{yellow}\hl{throat}}
    and {\sethlcolor{orange}\hl{wings}} (c) to reflect their identification features described on \href{https://www.allaboutbirds.org/guide/Eastern_Bluebird/id}{AllAboutBirds.org} \emph{(``Male Eastern Bluebirds are vivid, deep blue above and rusty or brick-red on the throat and breast'')}.
    After the edit, PEEB correctly predicts the input image into \textcolor{ForestGreen}{\class{Eastern Bluebird}} (softmax: 0.0445) out of 201 classes (c).
    That is, the dot product between the {\sethlcolor{orange}\hl{wings}} text descriptor and the same orange region increases from \textcolor{orange}{\footnotesize \sffamily 0.57} to \textcolor{orange}{\footnotesize \sffamily 0.74}.
    }
    \label{fig:change_descriptor}
\end{figure*}

For \textbf{birds} \bird, we first define the parts of interest for identifying a bird. 
We take the 15 parts defined in CUB \cite{wah2011caltech} and reduce them to 12 by merging similar parts, \eg \textit{left wing} and \textit{right wing} are merged into \textit{wings}. 
Using GPT-4 \citep{openai2023gpt4}, we construct a \emph{textual} descriptor 
to describe each bird part of every species (see \Cref{sec:descriptors}).
Next, \peeb localizes the 12 bird parts in the image and computes their matching scores with corresponding text descriptors (\cref{fig:change_descriptor}).
The sum of the 12 dot products between the paired visual and textual part embeddings would be the unnormalized distance (logits) between the input image and every class for classification
(\cref{fig:xclip_concept}).
For \textbf{dogs} \dog, we use a similar procedure.

To our knowledge, all existing public bird-image datasets (listed in \cref{tab:birdsoup}) are limited in size (less than 100K images per dataset) and in diversity (less than 1,500 species per dataset), impeding large-scale, vision-language, contrastive learning.
Therefore, for our pre-training, we construct \birdsoup, an exceptionally large dataset of bird images, comprising $\sim$290,000 images spanning across $\sim$11,000 species---essentially, \emph{all} known bird species on Earth (\cref{sec:dataset}).
\birdsoup is constructed from seven existing bird datasets and $\sim$55,000 new images that we collect from the \href{https://www.macaulaylibrary.org/the-internet-bird-collection-the-macaulay-library/}{Macaulay Library}.
Similarly, we build Dog-140, a large-scale dataset of 206K dog images.
Our main findings are:\footnote{Code \& data: \url{https://github.com/anguyen8/peeb}}

\begin{enumerate}

    \item CLIP-based classifiers rely mostly on class names in the prompt: The CUB accuracy of \mv model \cite{menon2023visual} drops drastically from 53.78\% to 5.89\% and 5.95\% after class names are removed or replaced by scientific names (\cref{sec:effect_on_descriptions}). 
    
    \item Our pre-trained \peeb outperforms CLIP-based classifiers by \increasenoparent{8} to \increasenoparent{29} percentage points (pp) in bird classification across \cub-200, \nabirds-555, and \inat-1486 (\cref{sec:clip_based_comparison}).

    \item \peeb allows defining new classes in text at test time (\cref{fig:change_descriptor}) without any further training. Besides explainability and editability, \peeb outperforms \emph{text concept-based} methods in both the generalized zero-shot (\cref{sec:concept_based_comparison}) and zero-shot setting (\cref{sec:real_zeroshot}).

    
    \item Compared with explainable CUB classifiers, \peeb scores an 88.80\% top-1 accuracy, on par with the best CUB-200 classifiers (81--87\% accuracy) that are trained via supervised learning and often \emph{not} editable (\cref{sec:target_finetune}).

    \item \peeb is applicable to multiple domains. 
    On Stanford Dogs-120, \peeb scores 92.20\%, substantially outperforming explainable models and on-par with SOTA black-box models (\cref{sec:applying_dogs}).

\end{enumerate}

\section{Related Work}
\label{sec:related_work}

\subsec{Ante- vs. post-hoc explanations}
It is common to build fine-grained classifiers based on CNNs \citep{he2016deep} or ViTs \citep{he2022transfg}.
Although high-performing, these models do not admit an \emph{ante}-hoc explanation interface \citep{gunning2021darpa} and therefore rely on \emph{post}-hoc interpretability methods, which tend to offer inaccurate and unstable, after-the-fact explanations \citep{rudin2019stop,bansal2020sam}.
PEEB's textual part-descriptors form an ante-hoc, natural-language explanation bottleneck that enables users to observe and edit the object attributes that contribute to each final prediction.
By editing text descriptors, users can re-program the model without any further re-training (\cref{fig:change_descriptor}).

\subsec{Prototypical Part Networks}
%
Like the explainable classifiers that learn part prototypes \citep{nauta2021neural,donnelly2022deformable,nauta2022looks,chen2019looks}, PEEB also operates at the object-part level.
However, there are two major differences.
First, the textual part descriptors in PEEB are human-understandable and editable.
In contrast, a part prototype \cite{chen2019looks} is not directly editable or interpretable to users, and often interpreted by showing the nearest training-set image patches.
Second, PEEB predicts 
a \emph{contextualized} embedding for each object part and its spatial information can be viewed by inputting to the Box MLP (see \cref{fig:xclip_concept}) for bounding-box visualization.



\subsec{Text-based Concept Bottlenecks}
Like PEEB, \citep{chen2020canzsl,zhu2018generative,rao2023dual,paz2020zest} also match visual part embeddings to text embeddings.
Yet, they (1) do not use CLIP and instead rely on TF-IDF text features; (2) require a trained bird-part detector to detect 7 bird parts.
In contrast, PEEB relies on CLIP, which admits easy text editability, and OWL-ViT, which serves as an open-vocabulary object-part detector that generalizes to many domains.

Recent vision-language models (VLMs) claim to be interpretable as they use textual concepts in the prompt.
Yet, some works that rely on class-wise differential captions \cite{esfandiarpoor2023follow} or learned concept weights \cite{yang2023language,panousis2023hierarchical,oikarinen2023labelfree,yuksekgonul2023posthoc} do \emph{not} generalize to unseen classes.
The most recent, similar work to PEEB might be LaBo \cite{yang2023language}, which; however, operates at the \emph{image} level instead of patch level, and does \emph{not} generalize to unseen classes.



Many CLIP-based classifiers \citep{han2023llms,pratt2022does,menon2023visual} rely heavily on having \emph{seen} class names in the prompt and thus are neither explainable nor editable to users.
Unlike CLIP-based models, \peeb reveals what image details are being used for classification by matching descriptors to corresponding visual object parts (\eg a bird's \textcolor{magenta}{\class{beak}} in \cref{fig:xclip_concept}).

\subsec{Attribute-based Classifiers} Attribute-Label Embedding (ALE) approaches \citep{akata2015label,yuksekgonul2023posthoc} employ a fixed set of attributes and train an attribute-to-label weight matrix for zero-shot classification. 
Several studies \citep{samuel2021generalized,xu2020attribute,hanouti2023learning} highlight its effectiveness on datasets like \cub, SUN \citep{5539970}, and AWA \citep{8413121}. 
Yet, in practice, ALE requires tabular data annotations for every new class in the dataset (e.g., 312 attributes per CUB species), editing the weight matrix, and model re-training.
In contrast, to add an unseen class to PEEB, users would only need to describe its 12 bird parts in natural language.

\section{Datasets}
\label{sec:dataset}


\subsection{Test classification benchmarks}
We test PEEB on three \bird bird classification datasets: CUB-200 \citeyearpar{wah2011caltech}, NABirds-v1 of 555 classes \citeyearpar{van2015building}, and iNaturalist \citeyearpar{van2021benchmarking} which has 1,486 bird classes.
For \dog dog images, we test PEEB on Stanford Dogs-120 \citeyearpar{khosla2011novel}.

\subsection{\birdsoup dataset construction}
\label{sec:bird11_construction}


We combine \emph{labeled} images from 7 distinct datasets and an extra $\sim$55K images (10,534 classes) from Cornell's Macaulay Library, to form a unified \textbf{\birdsoup} dataset\footnote{We do not redistribute the published datasets but release a script for reconstructing \birdsoup on \href{https://github.com/anguyen8/peeb}{Github}.} (\Cref{sec:birdsoup}) for large-scale pre-training.
To the best of our knowledge, \birdsoup, comprising 440,934 images spanning 11,183 classes, is the first bird dataset to encompass almost all species on Earth.
Since \peeb learns to match visual parts with textual descriptors, it requires that bird images be distinctly visible and sufficiently large for accurate part localization and matching (see \Cref{abl:effect_of_classes_and_images} for ablation studies).
However, small and ``hard-to-see'' bird images in \birdsoup make the dataset noisy and the training complex.
Thus, we harness \owlvitLarge \citep{minderer2022simple} to detect a bird in all images using the prompt \textit{``bird''} and filter out images where the detected bird's bounding box is smaller than $100 \times 100$ pixels.
We find OWL-ViT's bird detections to be fairly accurate---its mean Intersection over Union (IoU) between the predicted bird boxes and ground-truth boxes on CUB dataset is 0.91.

As class labels from different sources are either general (\eg ~\class{Cardinal}) or fine-grained (\eg ~\class{Yellow} vs. \class{Northern Cardinal}), we retain only the fine-grained species for more diverse training and exclude all general classes to avoid label ambiguity.
Following these filtering steps, the refined \birdsoup dataset retains 294,528 images across 10,811 classes (\cref{tab:birdsoup}).

For each species in \birdsoup, we generate a set of part-based descriptors using GPT-4 (\Cref{sec:descriptors}).
These generated descriptors (see \cref{fig:randomized_description}) may not be 100\% accurate but discriminative enough to help GPT-4V reach 69.40\% accuracy on the CUB-200 test set (Table~\ref{tab:concept_based}).
That is, in the same prompt, we feed each test image $x$ along with the 200 CUB classes' part-based descriptors and ask GPT-4V to select a matching class label for $x$ (details in \Cref{sec:descriptors_assessment}).
\begin{figure*}[t]
    \centering    \includegraphics[width=1\linewidth]{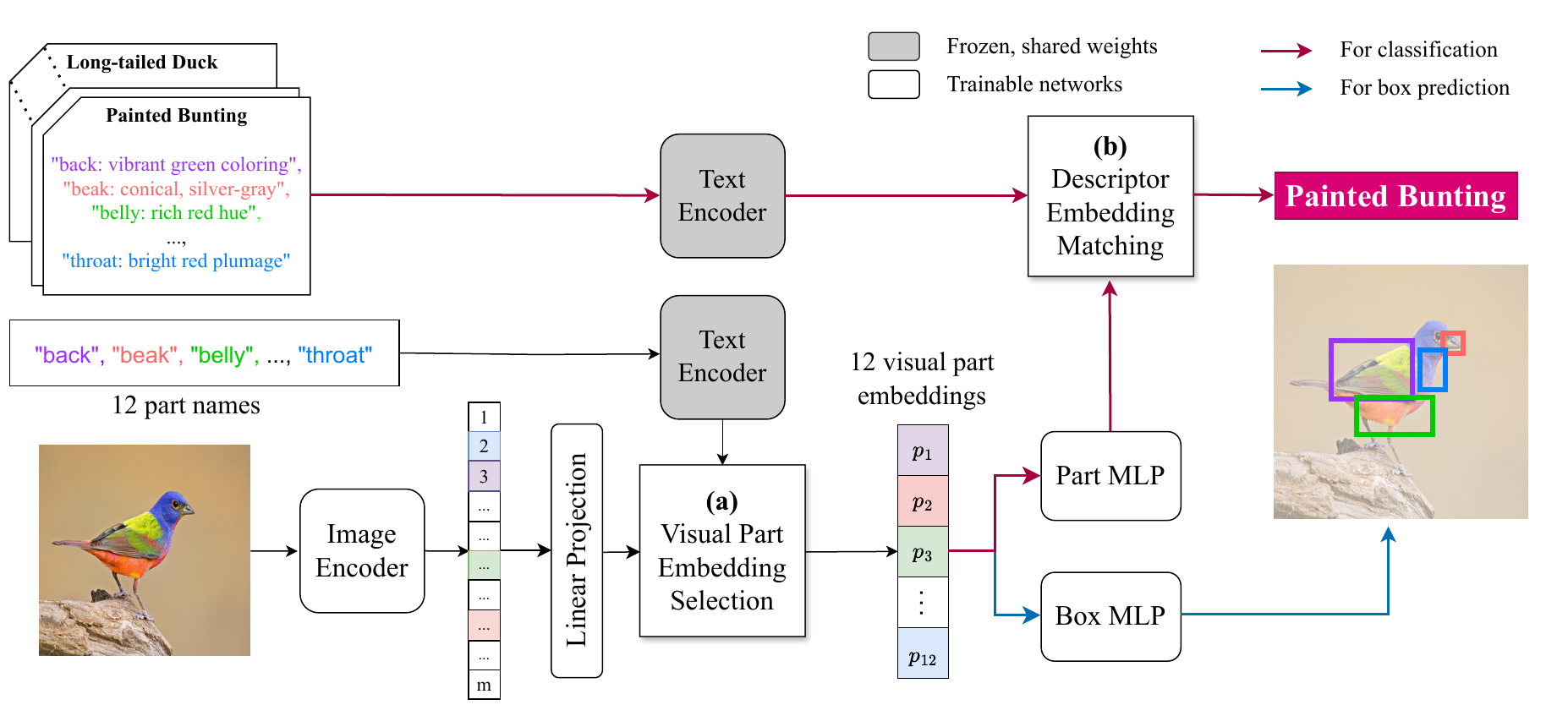}
    \caption{
    During inference, 12 visual part embeddings with the highest cosine similarity with encoded part names are selected \textbf{(a)}. 
    These visual part embeddings are then mapped (\textcolor{darkcyan}{$\longrightarrow$}) to bounding boxes via Box MLP. Simultaneously, the same embeddings are forwarded to the Part MLP and its outputs are then matched \textbf{(b)} with textual part descriptors to make classification predictions (\textcolor{darkmagenta}{$\longrightarrow$}). 
    \cref{fig:xclip_overview} shows a more detailed view of the same process.}
    \label{fig:xclip_concept}
\end{figure*}

\subsection{Dataset splits for contrastive pre-training}
\label{sec:data_splits}

There are two common settings in the zero-shot learning literature---standard zero-shot (\textbf{\zsl}) and generalized zero-shot (\textbf{\gzsl}).

\textbf{\zsl} is a stricter setup where a model is only tested on the \emph{classes} unseen during any prior training. 
We ensure test-set classes from datasets (\eg, \cub-200 or \nabirds-555) are not seen during pre-training. 
For example, to test on \cub under \zsl, we exclude all 200 \cub~classes and their images from our pre-training on Bird-11K.

Following the ZSL literature, we use the \cub split proposed by \citet{akata2015label} and two harder splits: Super-Category-Shared/Exclusive (SCS/SCE) by \citet{elhoseiny2017link}.
For example, in \zsl on \cub, we exclude all \cub classes in \birdsoup for pre-training and finetune only on the corresponding training set given by a \zsl split.

\textbf{\gzsl} is closer to the real-world setup where models are tested on both seen \& unseen classes.
CLIP's ``zero-shot'' tests technically fall under \gzsl as its Internet-scale training set might actually have images from the test classes.
To test PEEB under \gzsl, we exclude the \emph{test} sets of \cub, \nabirds, and \inat, and directly evaluate the Bird-11K-pretrained models without further finetuning.

\section{Method}
\label{sec:method}

\subsection{Backbone: \owlvit object-part detector} 
\label{model:owlvit}


\owlvit 
is an open-vocabulary detector that detects objects and parts in an image given a text prompt, even if the model is not explicitly finetuned to detect those concepts.
\owlvit consists of four networks (\cref{fig:xclip_concept}): 
(1) a ViT-based image encoder, (2) an architecturally identical text encoder, (3) a bounding-box regression head called Box MLP, and (4) and a Linear Projection.
Box MLP is a three-layer Multilayer Perceptron (MLP) with GELU activations \citep{hendrycks2016gaussian} after each of the first two layers. 
Linear Projection maps the visual and text embeddings to the same space (see Fig.~1 in \cite{minderer2022simple}).

\subsection{PEEB classifier}
\label{model:xclip}

\subsec{Architecture} 
\peeb (\cref{fig:xclip_concept}) has five networks: an image encoder, a text encoder, a Linear Projection, a Part MLP, and a Box MLP.

We introduce \textbf{Part MLP} to map the visual and textual part embeddings to the same space for computing dot products (logits) for classification (\textcolor{darkmagenta}{$\longrightarrow$} in \Cref{fig:xclip_concept}). 
This design allows PEEB to easily extend the number of classes without any re-training. 
Except for Part MLP, all components are adopted from the \owlvit framework. 
Details of all components are in \Cref{sec:model_details_apdx}.

\subsec{Inference} Given an input image, we first use the 12 generic part names to select the visual part embeddings based on cosine similarity.
These selected visual part embeddings are then simultaneously fed into both Part MLP and Box MLP. 

Box MLP predicts the bounding box from each part embedding.
We compute a dot product to measure the similarity between each embedding output from Part MLP and a corresponding part-descriptor embedding.
For classification, a class logit is the sum of the 12 dot products, which essentially computes the similarity between the 12 parts in the image and the 12 text descriptors of each class.

\subsection{Training strategy}
\label{sec:training_strategy}

\subsec{Trainable networks} In preliminary experiments, we find training only Part MLP (while keeping all other networks frozen) to result in poor accuracy.
Therefore, we train Part MLP from scratch and also finetune the image encoder, Linear Projection, and Box MLP.
We finetune all \owlvit components from their original weights.
In contrast, our proposed Part MLP starts from random weights.  
Our training has two phases: (a) 2-stage \textbf{pre-training} on the large-scale \birdsoup dataset and (b) \textbf{finetuning} on downstream tasks. 
More hyperparameter details are in \Cref{sec:model_details}.

\subsec{Objectives} We aim to train PEEB to classify images well while maintaining the ability to detect object parts.
This translates into \textbf{three training objectives}:
\textbf{(1)} Train the \text{Part MLP} contrastively using a symmetric cross-entropy (SCE) loss \citep{radford2021learning} to maximize the similarity between region-text pairs while minimizing the similarity for negative pairs; 
\textbf{(2)} Train the \text{Linear Projection} using a SCE loss to mimic \owlvit's behaviors (\ie the similarity matrix) for part selection; and 
\textbf{(3)} Train \text{Box MLP} to predict bounding boxes with DETR losses \citep{zheng2020end} \ie a linear combination of $\ell_1$ corner-to-corner distance loss and GIoU loss \citep{Rezatofighi_2019_CVPR}. 

All three losses are described in \Cref{apdx:objective_and_loss}.

\subsec{A challenge} when jointly minimizing all three losses above is that PEEB's validation loss improves significantly slowly perhaps because of some tension between the two SCE losses and the DETR detection loss.
To overcome this challenge, we split the pre-training phase into two stages: (1) first, train the image encoder and \text{Part MLP} for classification using the SCE loss; then (2) train the \text{Linear Projection} and \text{Box MLP} using the 2nd and 3rd loss so they can adapt their weights to the updated image encoder.
We always keep the \hlgray{text encoder} frozen since we want to preserve its generalizability to the descriptors of unseen objects.


\subsubsection{2-stage pre-training on \birdsoup}
\label{sec:pretraining}

\subsec{Stage 1: Contrastive learning} The image encoder and \text{Part MLP} are jointly trained using a SCE loss, which allows \peeb to learn to map the visual parts to corresponding text descriptors.
In this stage, we use a pre-trained \owlvitLarge to \emph{select} 12 part embeddings per input image (i.e., teacher forcing) to ensure the \emph{selection} of part embeddings is meaningful and consistent while the embeddings themselves are updating (see \cref{fig:pretraining_stage1}).

\subsec{Stage 2: Learning to detect from a teacher} 
%
After the image encoder is modified in Stage 1, we then train the Linear Projection and Box MLP jointly.
We use the \owlvitLarge as the teacher to train both components.
Using SCE loss, we train the Linear Projection such that the similarity matrix between the part-names and visual parts matches those of the teacher (\cref{fig:pretraining_stage2}, 1a--c, 2a--c).
Given the absence of human-annotated boxes for object parts, we train Box MLP to predict the same boxes as the predictions by \owlvitLarge using DETR losses (\cref{fig:pretraining_stage2}, 2d).
In this Stage 2, the image encoder is frozen while \text{Part MLP} is not involved.

\textbf{After 2-stage training}, \peeb can perform zero-shot classification and generate explanations. 

\subsubsection{Finetuning on classification tasks}
\label{sec:funetining}

We finetune the pre-trained PEEB on downstream tasks (\cub, \nabirds and \inat for birds and Dogs-120 for dogs) to further improve its accuracy. 
In this phase, to adapt to a downstream task, all components except the text encoder are trained jointly 
and the loss for \text{Part MLP} is changed from SCE (contrastive) to CE (classification) while the other two losses (DETR) are kept intact.

\section{Experiments \& Results}

We conduct extensive experiments to evaluate PEEB on multiple \bird bird datasets (\cub, \nabirds, \inat) and on \gzsl (\cref{sec:effect_on_descriptions,sec:concept_based_comparison}), \zsl (\cref{sec:real_zeroshot}) and also supervised learning settings.
We also find PEEB to perform well on \dog dog image classification on Dogs-120 (\cref{sec:applying_dogs}).

\begin{figure*}
    \centering
    
    \begin{subfigure}{\textwidth}
        {\scriptsize\makebox[0.8\textwidth][c]{Original Descriptors \textbf{(a)}}}%
        {\scriptsize\makebox[-0.05\textwidth][c]{\textcolor{red}{Randomized, wrong} descriptors \textbf{(b)}}}%
        
        {\tiny\makebox[0.3\textwidth][c]{\class{Blue Jay}}}%
        {\tiny\makebox[0.2\textwidth][c]{\textcolor{ForestGreen}{\class{Blue Jay}}~~0.0059}}%
        {\tiny\makebox[0.5\textwidth][c]{\textcolor{ForestGreen}{\class{Blue Jay}}~~0.0058}}%
        
        \begin{minipage}{0.02\textwidth}
            \rotatebox{90}{\scriptsize\mv \cite{menon2023visual}}
        \end{minipage}%
        \begin{minipage}{0.95\textwidth}
            \includegraphics[width=\linewidth]{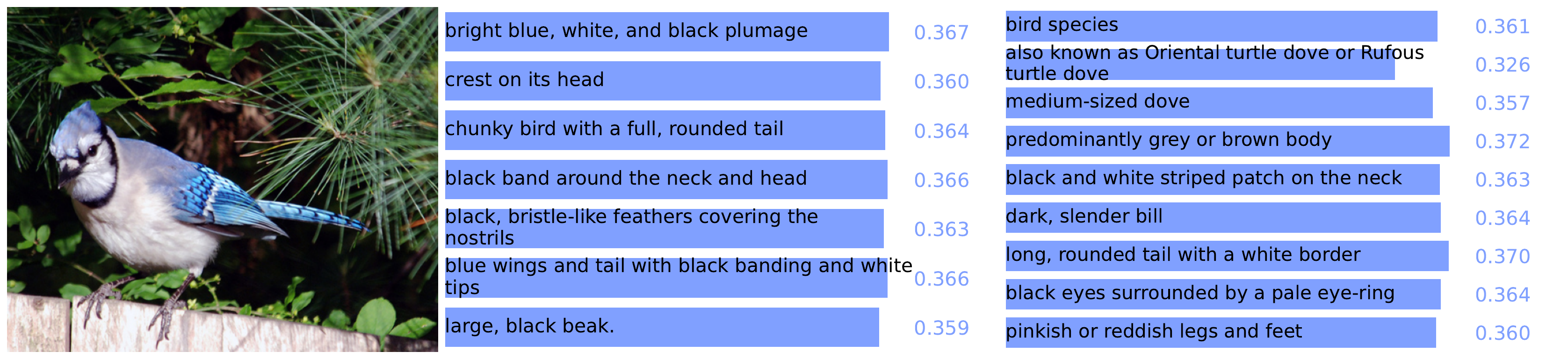}
        \end{minipage}
    \end{subfigure}
    
    \begin{subfigure}{\textwidth}
        \tiny\makebox[0.8\textwidth][c]{\textcolor{ForestGreen}{\class{Blue Jay}}~~0.6899 ~~\textbf{(c)}}%
        \tiny\makebox[-0.09\textwidth][c]{\textcolor{red}{\class{Least Tern}}~~0.0611 ~~\textbf{(d)}}%
        
        \begin{minipage}{0.02\textwidth}
            \rotatebox{90}{\scriptsize\peeb (\textbf{ours})}
        \end{minipage}%
        \begin{minipage}{0.95\textwidth}
            \includegraphics[width=\linewidth]{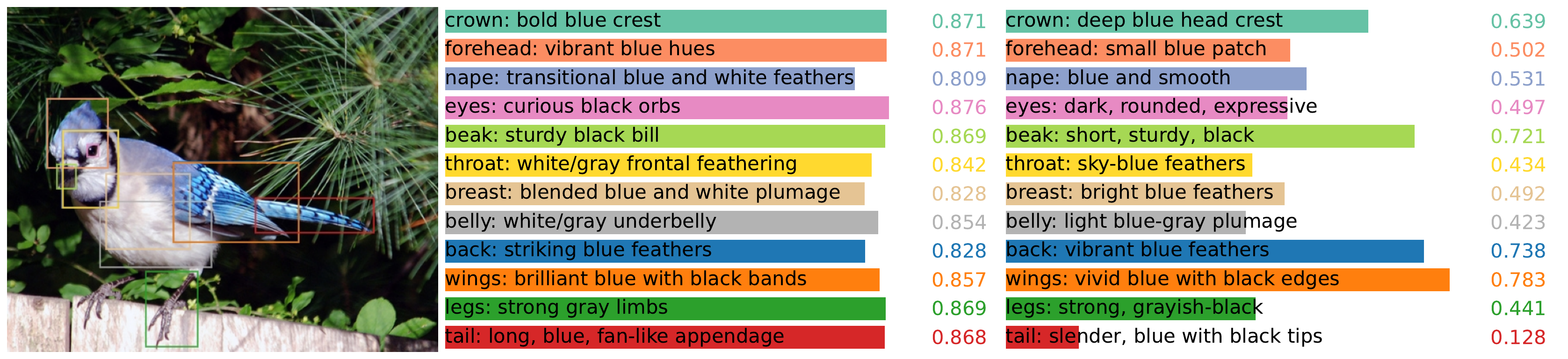}
        \end{minipage}
    \end{subfigure}
    
    \caption{With original descriptors, \mv \cite{menon2023visual} correctly classifies the input image into \textcolor{ForestGreen}{\class{Blue Jay}} \textbf{(a)}. 
    Yet, interestingly, when \textcolor{red}{randomly} swapping the descriptors of this class with those of other classes \textbf{(b)}, \mv's top-1 prediction remains unchanged, suggesting that the class names in the prompt (\eg, ``A photo of \texttt{\{class name\}}'') have the most influence over the prediction (not the expressive descriptors).
    In contrast, PEEB changes its top-1 prediction from \textcolor{ForestGreen}{\class{Blue Jay}} \textbf{(c)} to \textcolor{red}{\class{Least Tern}} \textbf{(d)} when the descriptors are \textcolor{red}{randomized}.
    }
    \label{fig:randomized_description}

\end{figure*}

\subsection{CLIP-based classifiers rely mostly on \class{\{class names\}} (\emph{not} descriptors)}
\label{sec:effect_on_descriptions}

\mv show that inserting extra GPT-3-generated descriptors into CLIP's prompts increases its accuracy on many tasks \cite{menon2023visual}.
Yet, it is unknown how important these expressive descriptors are compared to the class names. 
%
To answer this question, we conduct two experiments on all three models: CLIP, M\&V, and our PEEB.

\subsec{Experiment 1}
We evaluate the role of expressive descriptors to CLIP-based models and to PEEB by measuring the drop in CUB-200 accuracy of each model when the descriptors are randomized.

For M\&V and PEEB, we randomize the descriptors by swapping each descriptor with another from an arbitrary class (examples in \cref{fig:randomized_description}).

\subsec{Experiment 2} 
We test the dependence of models on class names by measuring the accuracy drop when they are replaced by scientific names (\eg, \class{Painted Bunting} $\to$ \class{Passerina ciris}) on \cub, \nabirds, and \inat. 


\subsec{Results}
When random descriptors are used, \mv accuracy drops marginally by \decreasenoparent{0.9} pp (\cref{tab:effect_on_descriptions}; 53.70\% $\to$ 52.88\%), showing that descriptors actually play a minimal role in model predictions.
Instead, CLIP and \mv mostly rely on class names (\eg, 53.78\% $\to$ 7.66\%; \cref{tab:clip_like_zeroshot})---their accuracy drops drastically when class names are replaced by scientific names, which are less common.

In contrast, the expressive part descriptors play a major role in PEEB whose accuracy decreases significantly to near random-chance (64.33\% $\to$ 0.88\%; \cref{tab:effect_on_descriptions}) when the descriptors are randomized. 
Indeed, in PEEB, the textual descriptors serve as editable and interpretable model parameters that can be refined and extended by humans to account for new classes
(\cref{fig:change_descriptor}).

\begin{table}[hbt]
\centering
\caption{Top-1 test accuracy (\%) on CUB-200 when using original, correct (a) vs. randomized, wrong descriptors (b).
See \cref{fig:randomized_description} for an example of the descriptors.
}
\label{tab:effect_on_descriptions}
\resizebox{0.5\textwidth}{!}{
\begin{tabular}{lcccl}\toprule
 & CLIP \citeyearpar{radford2021learning} & \multicolumn{2}{c}{\mv \citeyearpar{menon2023visual}} & \peeb \\
\cmidrule(l{2pt}r{2pt}){2-5}
 With class names & \greencheck & \greencheck & \redcross & ~~~~\redcross\\
\midrule
(a) Original descriptors & 52.02 & 53.78 & 5.89 & \textbf{64.33} \\
\midrule
(b) \textcolor{black}{Randomized} descriptors & \na & 52.88 & 0.59 & 0.88 \\
\bottomrule
\end{tabular}
}

\end{table}

\begin{table}[hbt]
\centering
\setlength\tabcolsep{4pt}
\caption{In the \textbf{\gzsl} setting, PEEB outperforms CLIP and \mv by a large margin, from \increasenoparent{8} to \increasenoparent{29} pp in top-1 accuracy (see \cref{sec:concept_based_comparison}).
PEEB is also $\sim$10$\times$ better than the other two models when class names are replaced by \textcolor{magenta}{scientific names}.
As PEEB does not use class names, its accuracy remains unchanged when class names are changed into the scientific ones.
}
\label{tab:clip_like_zeroshot}
\resizebox{0.47\textwidth}{!}{
\begin{tabular}{lcccccc}
\toprule
Acc (\%) & \multicolumn{2}{c}{\cub-200} & \multicolumn{2}{c}{\nabirds-555} & \multicolumn{2}{c}{iNaturalist-1486} \\
\cmidrule(l{2pt}r{2pt}){1-7}
CLIP \citeyearpar{radford2021learning} & 52.02 & \extraresults{5.95} & 39.35 & \extraresults{4.73} & 16.36 & \extraresults{2.03} \\
\cmidrule(l{2pt}r{2pt}){1-7}
\mv \citeyearpar{menon2023visual} & 53.78\tnote{*} & \extraresults{7.66} & 41.01 & \extraresults{6.27} & 17.57 & \extraresults{2.87} \\
\cmidrule(l{2pt}r{2pt}){1-7}
\peeb (\textbf{ours}) & \textbf{64.33} & (64.33) & \textbf{69.03} & (69.03) & \textbf{25.74} & (25.74) \\
\bottomrule
\end{tabular}
}
\end{table}

\subsection{Pre-trained \peeb outperforms CLIP-based classifiers in GZSL}
\label{sec:clip_based_comparison}

The dependence on class names (\cref{sec:effect_on_descriptions}) suggests that CLIP was exposed to these names during training.
Thus, for a fair comparison, we compare \peeb with CLIP-based classifiers in the \gzsl setting. 

\subsec{Experiment} We train \peeb on \birdsoup using the two-stage pre-training (described in \cref{sec:pretraining}), and then test it on \cub, \nabirds, and \inat without any finetuning.
That is, PEEB's contrastive pre-training is at the part level and therefore the model has not seen the species labels of images.

\subsec{Results} \peeb outperforms both CLIP and M\&V on all three datasets by huge margins of around \increasenoparent{10}, \increasenoparent{28}, and \increasenoparent{8} pp on \cub-200, \nabirds-555 and \inat-1486, respectively (see \cref{tab:clip_like_zeroshot}). 


\subsection{PEEB is superior to text descriptor-based classifiers in GZSL on CUB-200}
\label{sec:concept_based_comparison}

The advent of CLIP \citeyearpar{radford2021learning} by OpenAI enabled a class of image classifiers that match the input image with pre-defined textual prompts that may include class names or descriptors of the classes.
Yet, in contrast to PEEB, these descriptors often describe the entire image and are also matched (via dot product) with the entire image instead of image regions.
Here, we compare PEEB with these methods in the \gzsl setting on \cub-200.

\subsec{Experiment}
We repeat the same experiments in \cref{sec:clip_based_comparison}.
As these bird classifiers (listed in \cref{tab:concept_based}) were reported on CUB only (not \nabirds or \inat), our comparison is on CUB.

\subsec{Results}
\peeb exhibits superior \gzsl performance, outperforming recent text concept-based approaches by \increasenoparent{3} to \increasenoparent{10} pp (\cref{tab:concept_based}b). 
Compared to prior methods, \peeb is the only one to detect \emph{visual} object parts and match them with text descriptors.
Furthermore, attribute-based classifiers, \eg, \cite{yuksekgonul2023posthoc} require re-training to adapt to new classes or datasets (\eg, \nabirds or \inat) in the same domain.
In contrast, to apply \peeb to \nabirds or a new class, no training is required---it is necessary to only edit its text descriptors (see \cref{fig:change_descriptor}).
Interestingly, PEEB is 2nd-best model, only after GPT-4V (64.33\% vs. 69.40\%), which is given the same textual part descriptors as PEEB for all 200 CUB classes and asked to select a matching class for each image. 

\begin{table}[hbt]
\footnotesize
\centering
\caption{PEEB achieves SOTA CUB-200 accuracy among the \textbf{text descriptor-based} classifiers in \gzsl.\\ 
* \emph{1-shot learning.}~~~ $^{\dagger}$ \emph{$k$-means with} $k=32$.
}
\label{tab:concept_based}
\resizebox{0.5\textwidth}{!}{%
\begin{tabular}{llll}\toprule
Method & Acc (\%) & \texttt{\{c\}} & Textual descriptors \\
\midrule
\multicolumn{4}{l}{\cellcolor{gray!25} {\textbf{(a)} Vision-language models with class names \texttt{\{c\}} in the prompt}}\\
CLIP \citeyearpar{radford2021learning} & 52.02 & \redcheck & Image-level\\
M\&V \citeyearpar{menon2023visual} & 53.78 & \redcheck & Image-level\\
FuDD \citeyearpar{esfandiarpoor2023follow}   & 54.30 & \redcheck  & Image-level \\ 
\citet{han2023llms}                       & 56.13 & \redcheck & Image-level  \\
\midrule
\multicolumn{4}{l}{\cellcolor{gray!25} {\textbf{(b)} Vision-language models with text bottlenecks and \textcolor{ForestGreen}{\textbf{no}} class names \texttt{\{c\}}}}\\
LaBo \citeyearpar{yang2023language}            & 54.19$^{\dagger}$ & \greencross & Image-level \\ 
\citet{yan2023learning}                   & 60.27* & \greencross & Image-level, attribute-based \\
PEEB (\textbf{ours})                        & \textbf{64.33} & \greencross & Part-level\\
\midrule 
GPT-4V \citeyearpar{openai2023gpt4}   & 69.40   & \redcheck & Part-level\\ 
\midrule
\multicolumn{4}{l}{\cellcolor{gray!25} {\textbf{(c)} Concept-Bottleneck Models with attribute-based, non-textual bottlenecks}}\\
CBM \citeyearpar{koh2020concept}     & 62.90  &  \greencross & Attribute-based, tabular data \\ 
PCBM \citeyearpar{yuksekgonul2023posthoc}     & 61.00  & \greencross & Attribute-based, tabular data \\ 
\bottomrule
\end{tabular}
}
\end{table}

\subsection{\peeb generalizes to traditional \zsl}
\label{sec:real_zeroshot}

Since PEEB outperforms modern vision-language models in GZSL (\cref{sec:concept_based_comparison}), we are motivated to further compare PEEB with SOTA approaches in the traditional ZSL setting (where the test classes are excluded from all prior training).



\subsec{Experiment}
We evaluate PEEB on \textbf{two common \zsl splits}: (a) the CUB split \cite{akata2015label}; and (b) the Super-Category-Similar/Exclusive (SCS/SCE) splits \cite{elhoseiny2017link} on \cub and \nabirds. 
The SCS (Easy) and SCE (Hard) splits are designed to test two generalization levels (generalizing to close vs. distant unseen species).

Aligned with \zsl conventions, we exclude all species that exist in \cub or \nabirds from the pre-training and then finetune PEEB using the train/test splits by \citeauthor{akata2015label} and \citeauthor{elhoseiny2017link}. 
We randomly take $\sim$10\% of the training set as the validation set and choose the checkpoints based on the lowest validation loss.


\begin{table}[htp]
\centering
\setlength\tabcolsep{3.6pt}
\caption{
\peeb consistently outperforms other vision-language methods under Harmonic mean and especially in the hard split (SCE) by (\increasenoparent{5} to \increasenoparent{15}) points, highlighting its generalization capability on \zsl.
}
\label{tab:real_zeroshot}
\resizebox{0.48\textwidth}{!}{
\begin{tabular}{l|ccc|ccc}
\toprule
Methods & \multicolumn{3}{c|}{CUB} & \multicolumn{3}{c}{NABirds} \\
& Seen & Unseen & Mean  & Seen & Unseen & Mean \\
\midrule
\multicolumn{7}{c}{\textbf{(a)} \cellcolor{gray!25} Data split by \citet{akata2015label}}  \\
\midrule
\cloreClip \citeyearpar{han2022zero} & 65.80 & 39.10 & 49.05 & \multicolumn{3}{c}{\multirow{2}{*}{\na}} \\
\peeb (\textbf{ours}) & \textbf{80.78} & \textbf{41.74} & \textbf{55.04} & \\
\midrule
\multicolumn{7}{c}{\textbf{(b)} \cellcolor{gray!25} SCS/SCE splits by \citet{elhoseiny2017link}}  \\
\midrule
& SCS & SCE & Mean  & SCS & SCE & Mean \\
& (Easy) & (Hard) & & (Easy) & (Hard) & \\
\midrule
\ssgadet \citeyearpar{ji2018stacked} & 42.90 & 10.90 & 17.38 & 39.40 & 9.70 & 15.56\\
\grzsl \citeyearpar{zhu2018generative} & 44.08 & 14.46 & 21.77 & 36.36 & 9.04 & 14.48  \\
ZEST \citeyearpar{paz2020zest}   & \textbf{48.57} & 15.26  & 23.22 & 38.51 & 10.23 & 16.17 \\
CANZSL \citeyearpar{chen2020canzsl}      &  45.80  & 14.30   & 21.12  &  38.10 & 8.90 & 14.43 \\ 
\dgrzsl \citeyearpar{kousha2021zero} & 45.48 & 14.29 & 21.75 & 37.62 & 8.91 & 14.41\\ 	
DPZSL \citeyearpar{rao2023dual}         & 45.40 & 15.50   &  23.11 & \textbf{40.80} & 8.20 & 13.66\\ 
\peeb (\textbf{ours})& 44.66 & \textbf{20.31} & \textbf{27.92} & 28.26 & \textbf{24.34} & \textbf{26.15}\\
\bottomrule
\end{tabular}
}
\end{table}

\subsec{Results}
By a large margin, PEEB outperforms \cloreClip, a SOTA CUB method in the \citeyearpar{akata2015label} split, on both seen and unseen classes (\cref{tab:real_zeroshot}a).
On the \citeyearpar{elhoseiny2017link} splits, PEEB is the SOTA in the Hard set on both \cub and \nabirds datasets (\cref{tab:real_zeroshot}b).
That is, PEEB is better in generalizing to distant, unseen classes.
This may be because PEEB decomposes both the image and the text descriptors into part-level features, which can re-combine to match an arbitrary unseen class (as illustrated in \cref{fig:change_descriptor}).

Interestingly, on both \cub and \nabirds, PEEB is competitive but \emph{not} SOTA on the Easy sets (\cref{tab:real_zeroshot}b; Easy)---those classes that are close to the training-set classes and thus considered easier to identify.
Overall, considering the harmonic mean over both Easy and Hard accuracy scores, PEEB is the SOTA on both \cub and \nabirds.



\subsection{Finetuning the pre-trained PEEB on CUB-200 yields a competitive explainable classifier in supervised learning}
\label{sec:target_finetune}

After finding that PEEB performs well in both \gzsl (\cref{sec:concept_based_comparison}) and \zsl settings (\cref{sec:real_zeroshot}), here we test finetuning the pre-trained PEEB on CUB-200. 
That is, we compare PEEB against SOTA explainable classifiers in the supervised learning setting to gain insights into our method's adaptability to downstream tasks.


\subsec{Experiment}
To understand the impact of pre-training and image resolution, we test finetuning three different PEEB variants: (1) PEEB initialized from \owlvitBase \textbf{without pre-training} on \birdsoup; (2) PEEB initialized from \owlvitBase with pre-training (described in \cref{sec:clip_based_comparison}); and (3) PEEB initialized from \owlvitBaseSixteen with pre-training.
We take each PEEB model and finetune \emph{all} components on \cub-200,
for 30 epochs with a batch size of 30, a learning rate of $2\times10^{-5}$.
Detailed hyperparameters are in \cref{tab:fintuned_models}.

\subsec{Results}
Without pre-training, PEEB reaches 77.80\% top-1 accuracy on CUB-200.
Yet, first pre-training on \birdsoup and then finetuning on \cub yields 86.73\%, the best among all \emph{explainable} classifiers (\cref{tab:exp_finetune}b--c).
Besides, pre-training PEEB from the higher-resolution \owlvitBaseSixteen results in a further gain of \increasenoparent{2.07} (86.73\% $\to$ 88.80\%), which is intuitive since fine-grained classification is known to benefit from higher resolutions.

For a complete assessment, we compare and find PEEB to underperform SOTA standard, black-box classifiers by a few points (\cref{tab:exp_finetune}a).


\begin{table}[hbt]
\centering
\setlength\tabcolsep{1.5pt}
\caption{PEEB is a state-of-the-art, explainable CUB-200 \bird classifiers in the \textbf{supervised} learning. 
}
\label{tab:exp_finetune}
\resizebox{0.48\textwidth}{!}{
\begin{tabular}{lrll}\toprule
Methods & Model size & Backbone & Acc (\%)  \\
\midrule
\multicolumn{4}{l}{\cellcolor{gray!25} \textbf{(a)} SOTA \textbf{black}-box classifiers}  \\
Base (ViT) \citeyearpar{touvron2021training} & 22M & DeiT-S \citeyearpar{touvron2021training} & 84.28 \\
ViT-Net \citeyearpar{kim2022vit} & 26M & DeiT-S & 90.10 \\
\midrule
\multicolumn{4}{l}{\cellcolor{gray!25} \textbf{(b)} Concept-bottleneck classifiers}  \\
CBM \cite{koh2020concept} & 11M & ResNet-18 & 80.10 \\
CPM \cite{panousis2023hierarchical} &  155M & ViT-B/16 & 72.00 \\
CDM \cite{oikarinen2023labelfree} &  155M & ViT-B/16 & 74.31\\
LaBo \cite{yang2023language} & 427M & ViT-L/14 & 81.90 \\
\midrule
\multicolumn{4}{l}{\cellcolor{gray!25} \textbf{(c)} Part-based, explainable classifiers}  \\
ProtoPNet \citeyearpar{chen2019looks} & 22M & DeiT-S & 84.04  \\
ProtoTree \citeyearpar{nauta2021neural} & 92M & ResNet-50 & 82.20 \\
TesNet \citeyearpar{wang2021interpretable} & 79M & DenseNet-121 & 84.80 \\
Deformable ProtoPNet \citeyearpar{donnelly2022deformable} & 23M & ResNet-50 & 86.40  \\ 
ProtoPFormer \citeyearpar{xue2022protopformer} & 22M & DeiT-S & 84.85 \\

\peeb (\textbf{ours}) & 155M &  \\
~~~\textcolor{lightgray}{\st{pre-training +}} finetuning \emph{only} & 155M & \owlvitBase & {77.80}  \\
~~~pre-training + finetuning & 155M & \owlvitBase & \textbf{86.73}  \\
~~~pre-training + finetuning & 155M & \owlvitBaseSixteen & \textbf{88.80}  \\
\bottomrule
\end{tabular}
}
\end{table}

\subsection{Applying \peeb~to \dog dog identification}
\label{sec:applying_dogs}

\begin{figure*}[t]
    \centering
    \includegraphics[width=1\linewidth]{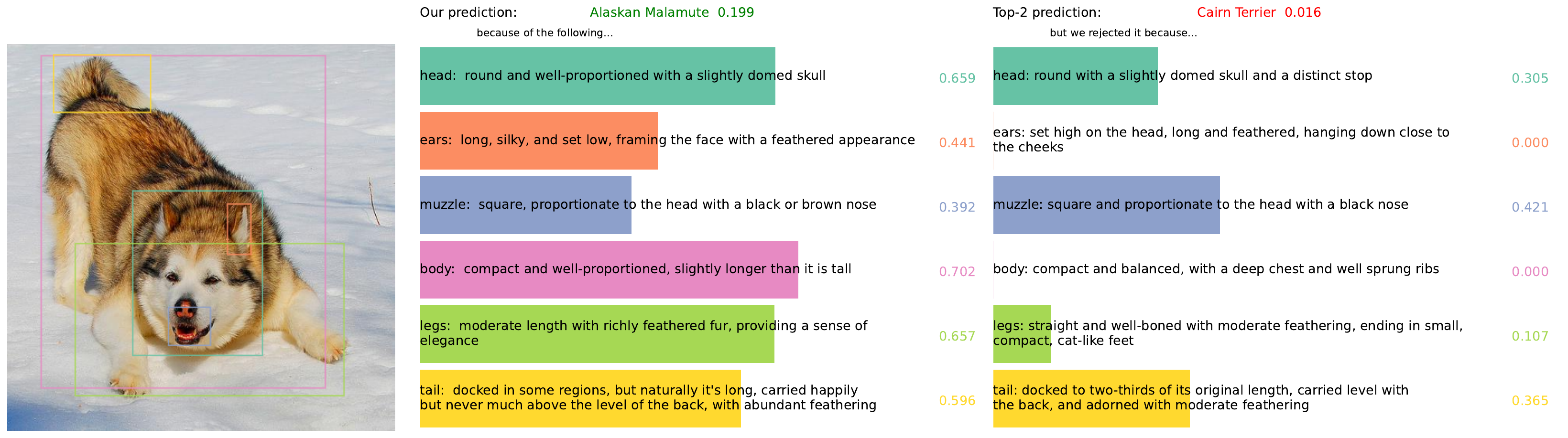}
    \caption{\peeb classifies this Dogs-120 image into \textcolor{ForestGreen}{\class{Alaskan Malamute}} (softmax: \textcolor{ForestGreen}{\footnotesize \sffamily 0.199}) due to the matching between the image regions and associated textual part descriptors.
    In contrast, the explanation shows that the input image is not classified into \class{Cairn Terrier} mostly because its \class{ears} and \class{body} regions do \emph{not} match the text descriptors, \ie, dot products are 
    \textcolor{orange}{\footnotesize \sffamily 0.000}
    and 
    \textcolor{magenta!90!black}{\footnotesize \sffamily 0.000}, respectively.
    See \Cref{sec:qualitative_examples} for more qualitative examples.
    }
    \label{fig:sd_explaination}
\end{figure*}


We have found that our pre-training dataset construction and PEEB performs well for bird identification.
By design, our method is \emph{not} specific to birds but is instead applicable to any fine-grained classification domains assuming that the object is decomposable into parts.
Here, we show that our method performs well on dog image classification as well.

\subsec{Pre-training dataset construction}
First, we define a set of six dog parts that humans use to identify dog species.
We use all 4 dog parts defined by PartImageNet \cite{he2021partimagenet}---\textit{head, body, legs,} and \textit{tail}---and two more parts---\textit{muzzle} and \textit{ears}---based on our manual image examination.

We combine ImageNet-21K and Stanford Dogs-120 into Dog-140, our large-scale pre-training dataset spanning 140 dog species (details in \cref{sec:dogsoup}).
For each class, we prompt GPT-4 to get the descriptors for 6 parts.
For each image in Dog-140, we run \owlvitLarge to detect the corresponding boxes for 6 pre-defined parts.

\subsec{Experiment}
Following the supervised learning experiment in \cref{sec:target_finetune}, we first we pre-train \peeb (initialized from \owlvitBase) on \dogsoup and then further finetune it on Dogs-120.

\subsec{Results} 
Finetuning PEEB on Dogs-120 from \owlvitBase without pre-training on Dog-140 results in a 74.17\% top-1 accuracy on Dogs-120 (\cref{tab:exp_pretrain_finetune_dog}b).
In contrast, pre-training on Dog-140 only without finetuning results in much better Dogs-120 accuracy of 87.38\%.
That is, our contrastive pre-training helps model generalize (in a \gzsl setting) while directly finetuning on Dogs-120 perhaps yields an overfitting model.
Yet, pre-training and then finetuning reaches the best supervised learning accuracy of 92.20\%, which is SOTA among all explainable models reported on Dogs-120.

Besides, PEEB offers novel, editable image-text grounding explanations (see \cref{fig:sd_explaination}).

\begin{table}[hbt]
\centering
\setlength\tabcolsep{1.5pt}
\caption{In the \textbf{supervised} learning setting, PEEB is the state-of-the-art explainable, Stanford Dogs-120 \dog classifiers and competitive w.r.t. SOTA black-box models.}
\label{tab:exp_pretrain_finetune_dog}
\resizebox{0.48\textwidth}{!}{
\begin{tabular}{lrlc}\toprule
Methods & Model size & Backbone & Acc (\%)  \\
\midrule
\multicolumn{4}{l}{\cellcolor{gray!25} \textbf{(a)} SOTA \textbf{black}-box classifiers}  \\
TransFG \citeyearpar{he2022transfg} & 86M & ViT-B/16 & 92.30  \\
ViT-Net \citeyearpar{pmlr-v162-kim22g} & 86M & DeiT-B & 93.60  \\
SR-GNN \citeyearpar{srgnn} & 32M & Xception & 97.00  \\
\midrule
\multicolumn{4}{l}{\cellcolor{gray!25} \textbf{(b)} Explainable methods}  \\
FCAN \citeyearpar{Liu2016FullyCA} & 50M & ResNet-50 & 84.20 \\
RA-CNN \citeyearpar{Fu_2017_CVPR} & 144M & VGG-19 & 87.30 \\
ProtoPNet \citeyearpar{chen2019looks} & 22M & DeiT-S & 77.30  \\
Deformable ProtoPNet \citeyearpar{donnelly2022deformable} & 23M & ResNet-50 & 86.50  \\ 
\peeb (\textbf{ours}) & 155M &  \\
~~~\textcolor{lightgray}{\st{pre-training +}} finetuning \emph{only} & 155M & \owlvitBase & {74.17}  \\
~~~pre-training \textcolor{lightgray}{\st{+ finetuning}} & 155M & \owlvitBase & {87.37}  \\
~~~pre-training + finetuning & 155M & \owlvitBaseSixteen & \textbf{92.20}  \\
\bottomrule
\end{tabular}
}
\end{table}

\section{Discussion and Conclusion}
We introduce \peeb, a unique, novel explainable classifier due to its editability (\cref{fig:change_descriptor}) and operation at the part level on both image and text sides.
The part-level operation makes \peeb applicable to fine-grained classification.
Yet, it is also interesting to extend \peeb into an object-level model for multi-domain tasks like ImageNet or VQA.

Besides enabling users to edit PEEB's text descriptors to re-program PEEB, it might also be promising to let users edit the bounding boxes while working with \peeb to improve the human-AI team accuracy \cite{nguyen2024allowing}.
On object detection, PEEB's Box MLP performs on-par with \owlvitBase based on quantitative (\Cref{sec:box_evaluation}) and qualitative results  (\Cref{sec:qualitative_examples}).




Finally, we contribute to the broader research community by curating the \birdsoup and \dogsoup datasets and showing that it is possible to leverage them for large-scale training.

\section{Limitations}
\label{sec:limitations}

\paragraph{Text encoder may not fully comprehend the bird descriptors} Our CLIP text encoder, pre-trained on an Internet-scale image-text dataset \cite{radford2021learning}, may not fully capture the intricate details specific to birds. 
Furthermore, the CLIP text encoder is known to suffer from the \emph{binding} problem and do not understand some logical operators such as ``and'', ``or'', or negation.
PEEB accuracy depends heavily on the quality of the text encoder.

\paragraph{Assumption that object parts mostly visible} 
PEEB operates based on the assumption that most if not all of the object parts are visible in the image.
In cases where a part is missing or occluded, the model may still assign a non-zero similarity score (i.e. a non-zero dot product between the image-part embedding and its associated text descriptor), which makes it harder to separate classes.
It might be beneficial to incorporate extra training samples and specifically encourages PEEB to assign \emph{zero} image-text similarity score to the missing or occluded parts.

\paragraph{Hallucinations in GPT-4 descriptors} The accuracy of PEEB is directly governed by the accuracy of descriptors, which are currently generated by GPT-4. 
Yet, our manual assessment over 20 bird classes reveals that, on average, 45\% of these descriptors do not accurately reflect the birds' features (\Cref{sec:gpt_noise_measure}). Also, we observe that revising certain descriptors in the \cub dataset led to a \textbf{significant} improvement of \increasenoparent{10} points in classification accuracy for those classes (\Cref{sec:revise_description}). This primitive observation suggests that \peeb can be further improved if trained with more accurate, human-labeled descriptors.


\section*{Acknowledgement}
We are grateful to Cornell's Macaulay Library for providing us with $\sim$55K images categorized into 10,534 bird species for the large-scale pre-training.
We thank Pooyan Rahmanzadehgervi, Giang Nguyen, Tin Nguyen, and Hung Huy Nguyen from Auburn University for their helpful feedback on our early results.
We also thank Phat Nguyen for her valuable support and feedback. 
AN is supported by NaphCare Foundations, Adobe gifts, and NSF grant no. 2145767.

\clearpage
\bibliography{references}

\clearpage

\onecolumn

    
    


\newcommand{\beginsupplementary}{%
    \setcounter{table}{0}
    \renewcommand{\thetable}{A\arabic{table}}%
    
    \setcounter{figure}{0}
    \renewcommand{\thefigure}{A\arabic{figure}}%
    
    \setcounter{section}{0}
    \renewcommand{\thesection}{\arabic{section}}
}

\beginsupplementary%
\appendix

\newcommand{\toptitlebar}{
    \hrule height 4pt
    \vskip 0.25in
    \vskip -\parskip%
}
\newcommand{\bottomtitlebar}{
    \vskip 0.29in
    \vskip -\parskip%
    \hrule height 1pt
    \vskip 0.09in%
}

\newcommand{\suptitle}{Appendix for:\\\papertitle}

\newcommand{\maketitlesupp}{
    \newpage
    \onecolumn
        \null
        \vskip .375in
        \begin{center}
            \toptitlebar
            {\Large \bf \suptitle\par}
            \bottomtitlebar
            \vspace*{24pt}
            {
                \large
                \lineskip=.5em
                \par
            }
            \vskip .5em
            \vspace*{12pt}
        \end{center}
}

\maketitlesupp%

\section{Architecture details}
\label{sec:model_details_apdx}

\subsection{Image encoder and text encoder}
We employ the image encoder and text encoder from \owlvit. In order to maintain a general understanding of natural languages and avoid overfitting our training samples, we keep the text encoder frozen for all training and experiments. This setup allows our design to be flexible about the choice of text encoder, e.g., one can easily replace the text encoder without changing other architecture.

\subsection{Linear projection (for part embedding selection)}
The image embedding will be forwarded to a \textbf{Linear Projection} layer (\href{https://github.com/anguyen8/peeb/blob/0ae217336de95e9bb70d33c3b7161e2eea834172/src/owlvit_cls.py#L27C14-L27C20}{see detail implementation here}), which is composed of a learnable logit scale, a learnable logit shift, and an Exponential Linear Unit (ELU) activation function. These processed image embeddings then have the same dimension as the text embeddings. For \owlvitBase, the image embeddings are projected from 768 to 512. We select a single image embedding for each text query. In this context, the text queries correspond to the component names of the target object, which includes twelve distinct parts. This selection is based on the cosine similarity between the projected image embeddings and the text embeddings. Finally, the chosen images embeddings (before projection) will be sent to the \textbf{Part MLP} for classification and \textbf{Box MLP} for box prediction (\cref{fig:xclip_overview}, Step 1).


\subsection{Part MLP}
We introduce \textbf{Part MLP} to enable part-based classification (\href{https://github.com/anguyen8/peeb/blob/0ae217336de95e9bb70d33c3b7161e2eea834172/src/owlvit_cls.py#L117}{see implementation detail here}). 
It comprises a three-layer MLP with GELU activations \citep{hendrycks2016gaussian} . 
\textbf{Part MLP} takes in the selected part embeddings (\ie~output of step 1 in \cref{fig:xclip_overview}) and  outputs a vector of size $\mathbb{R}^{d}$ for each part, where $d$ is the dimension of descriptor embeddings (for \owlvitBase, the input dimension is 768, and $d=512$).
Part MLP is trained to map the selected part embeddings to the same dimensional space with descriptor embeddings to compute final logits for classification.




\subsection{Box MLP} 
The \textbf{Box MLP} retained from \owlvit consists of a three-layer MLP (\href{https://github.com/anguyen8/peeb/blob/0ae217336de95e9bb70d33c3b7161e2eea834172/src/owlvit_cls.py#L69}{see here for implementation detail}). It takes the visual embedding as input and generates a four-element vector corresponding to the center coordinates and size of a bounding box (e.g., \texttt{[x, y, width, height]}). It is important to note that the image embedding inputs of \textbf{Box MLP} and \textbf{Part MLP} layers are the same, as shown in \cref{fig:xclip_overview}, Step 2.


\subsection{Visual part embedding selection}
\begin{figure*}[!hbt]
    \centering    
    \includegraphics[width=1\linewidth]{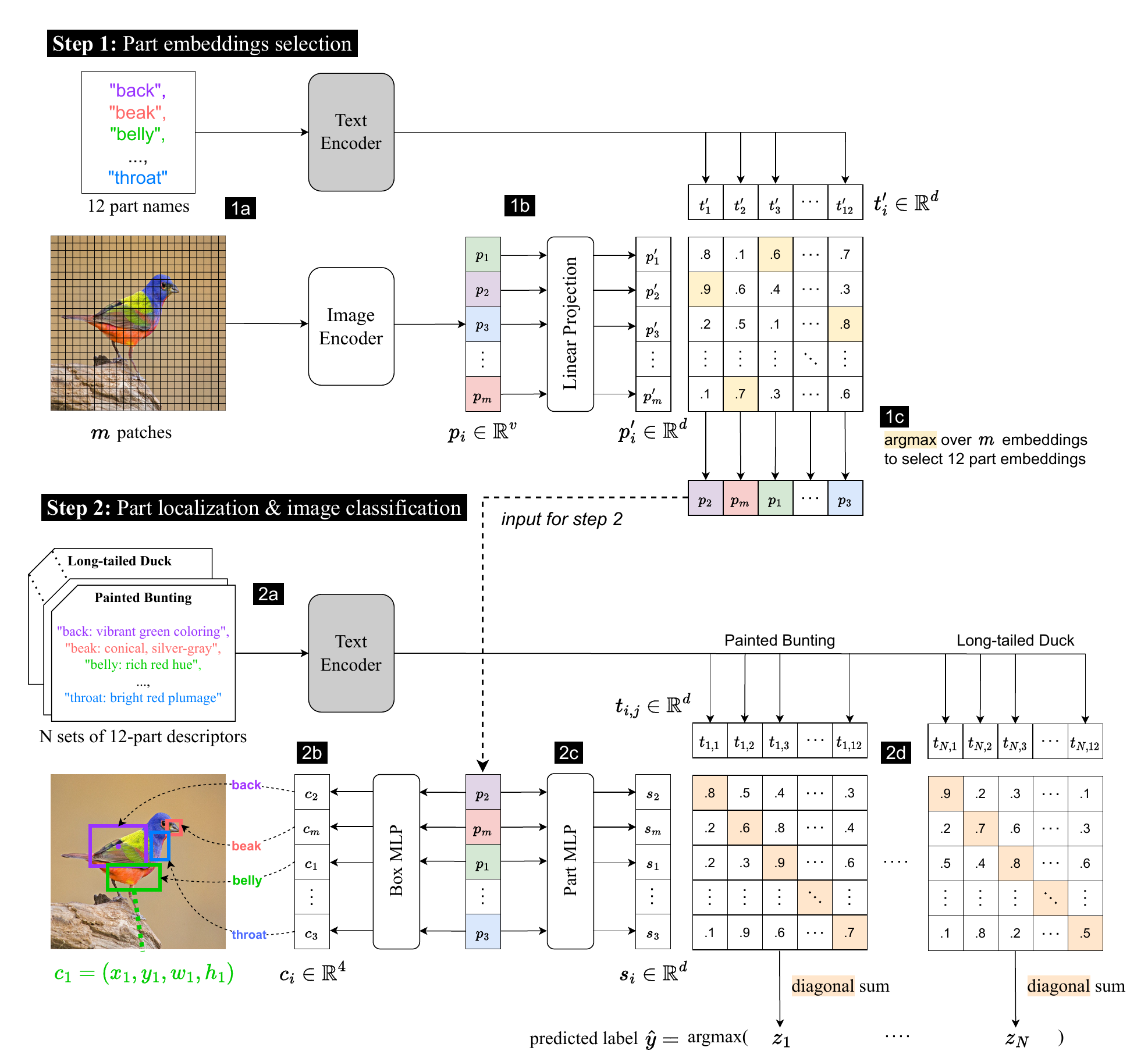}
    \caption{
    During the test time using PEEB, we perform 2 steps.\\
    \textbf{Step 1}: 
    (a) Encode an input image and texts (\ie 12 part names) by the image and text encoder to get patch embeddings \textbf{$p_i$} and text embeddings $t'_i$.
    (b) Feed $p_i$ to \text{Linear Projection} to get $p'_i$ in the same dimensional space with $t'_i$ and compute dot product between $\{p'_i\}$ and $\{t'_i\}$.
    (c) $\argmax$ over $m$ embeddings to select 12 part embeddings.
    \\\textbf{Step 2}:
    (a) Encode input texts (\ie N sets of 12-part descriptors) with the same text encoder to get $t_i$.
    (b) Feed the selected part embeddings to \text{Box MLP} to localize parts (in center format).
    (c) Also feed the selected part embeddings to \text{Part MLP} to get $s_i$ in the same dimensional space with $t_i$
    (d) Compute a dot product between $\{s_i\}$ and $\{t_i\}$, then diagonal sum for each class and $\argmax$ over logits to get predicted label $\hat{y}$. 
    }
    \label{fig:xclip_overview}
\end{figure*}

As shown in \cref{fig:xclip_overview} step 1, 1c, the image embeddings are first projected by a Linear Projection layer and compute the dot product with the encoded part names. The image embeddings (before Linear Projection) are chosen as visual part embeddings by selecting the embedding that has the highest similarity scores with the corresponding part after the Linear Projection.

\newpage
\clearpage

\subsection{Descriptor embedding matching} 
To enhance the model's flexibility, we do not use a linear layer for classification. Instead, we adopt a strategy similar to CLIP: we compute the similarity matrix of the projected visual embeddings (image embeddings after processing by the \textbf{Part MLP}) and the text embeddings. Then, we sum the corresponding similarities of each part in the class; the class with the highest score is considered the predicted class as shown in \cref{fig:xclip_overview}, step 2, 2d. This design enables our proposed method to perform arbitrary ways of classification.

\subsection{Implementation details}
Our experiments are conducted under PyTorch \cite{NEURIPS2019_9015}. We employ HuggingFace's \cite{wolf-etal-2020-transformers} implementation of \owlvit and use their pre-trained models. The DETR losses implementation \cite{carion2020end} is employed directly from their official implementation. 

\subsection{Training hyperparameters}
\label{sec:model_details}

We provide the hyperparameters of all models trained in this work. 
\Cref{tab:preretraining_models} shows the details of the pre-training models. \Cref{tab:fintuned_models} presents the details of the finetuned models. All trainings utilize optimizer AdamW with Plateau Scheduler.

\subsection{Computational budget and infrastructures}
We use 8 Nvidia RTX A100 GPUs for our experiments. The pertaining approximate takes $\sim$24 hours on \birdsoup. The finetuning takes 2 to 4 hours with one single GPU.

\subsection{Pre-training and Finetuning objectives}
\label{apdx:objective_and_loss}
\begin{figure*}[hbt!]
    \centering
    \includegraphics[width=1\linewidth]{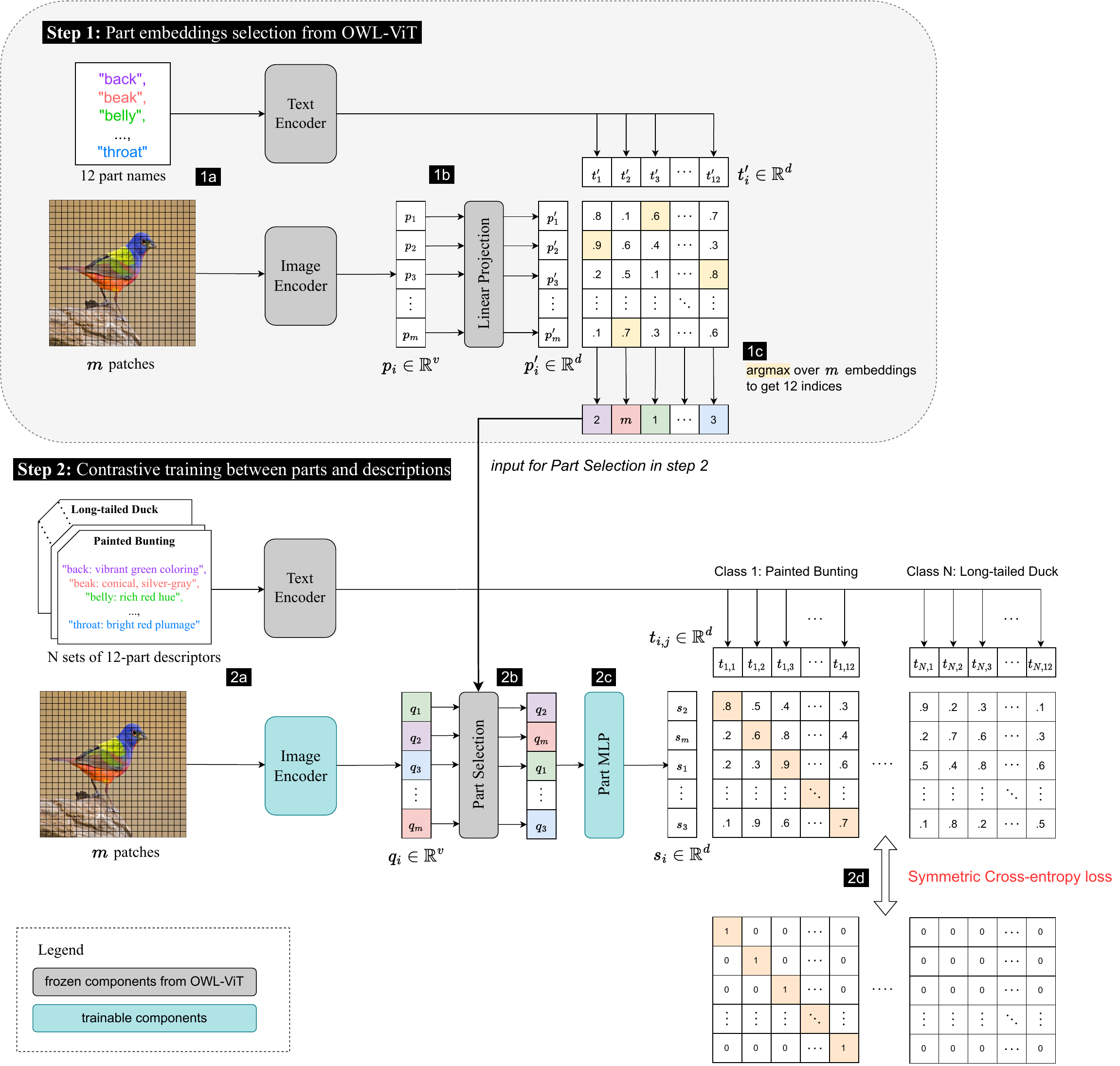}
    \caption{In pre-training stage 1, the objective is to let the Image Encoder learn the general representation of different parts of the birds. Therefore, in pre-training stage 1, we train the \textcolor{TealBlue}{\textit{Image Encoder}} and \textcolor{TealBlue}{\text{Part MLP}} contrastively. During the training, the \textbf{Step 1} utilizes a teacher model (\owlvitBase) to help \peeb select 12 part embeddings. In \textbf{Step 2}, we update the model with symmetric Cross-Entropy loss. Here's the flow of \textbf{Step 1}: (1a) We utilize the teacher model to encode 12 part names and the image to derive the text embedding $\vt'_i$, and the patch embedding $\vp_{i}$. (1b) Then the patch embeddings $\vp$ is forwarded to \text{Linear Projection} to obtain $\vp'$, matching the dimension of $\vt'$. (1c) We compute the dot product between $\vp$ and $\vt'$ and apply $argmax$ over $\vp$ to derive 12 indices. In \textbf{Step 2}: (2a), We first encode the descriptors and the image with the \textit{Text Encoder} and \textcolor{TealBlue}{\textit{Image Encoder}} to obtain descriptor embeddings $\vt$ and patch embeddings $\vq$. (2b), Then we select the 12 patch embeddings based on the 12 indices from (1c). (2c), The 12 patch embeddings then forwarded to \textcolor{TealBlue}{\text{Part MLP}} to derive $\vs$, which has the same dimension as $\vt$. Then, we compute the similarity matrix for the patch embedding and the descriptor embedding by computing the dot product between $\vs$ and $\vt$. (2d), we construct a one-hot encoded matrix based on the descriptors as the ground truth label and minimize the Symmetric Cross-Entropy loss between the similarity matrix in (2c) and the ground truth label.}
    \label{fig:pretraining_stage1}
\end{figure*}

\begin{figure*}[htb!]
    \centering
    \includegraphics[width=1\linewidth]{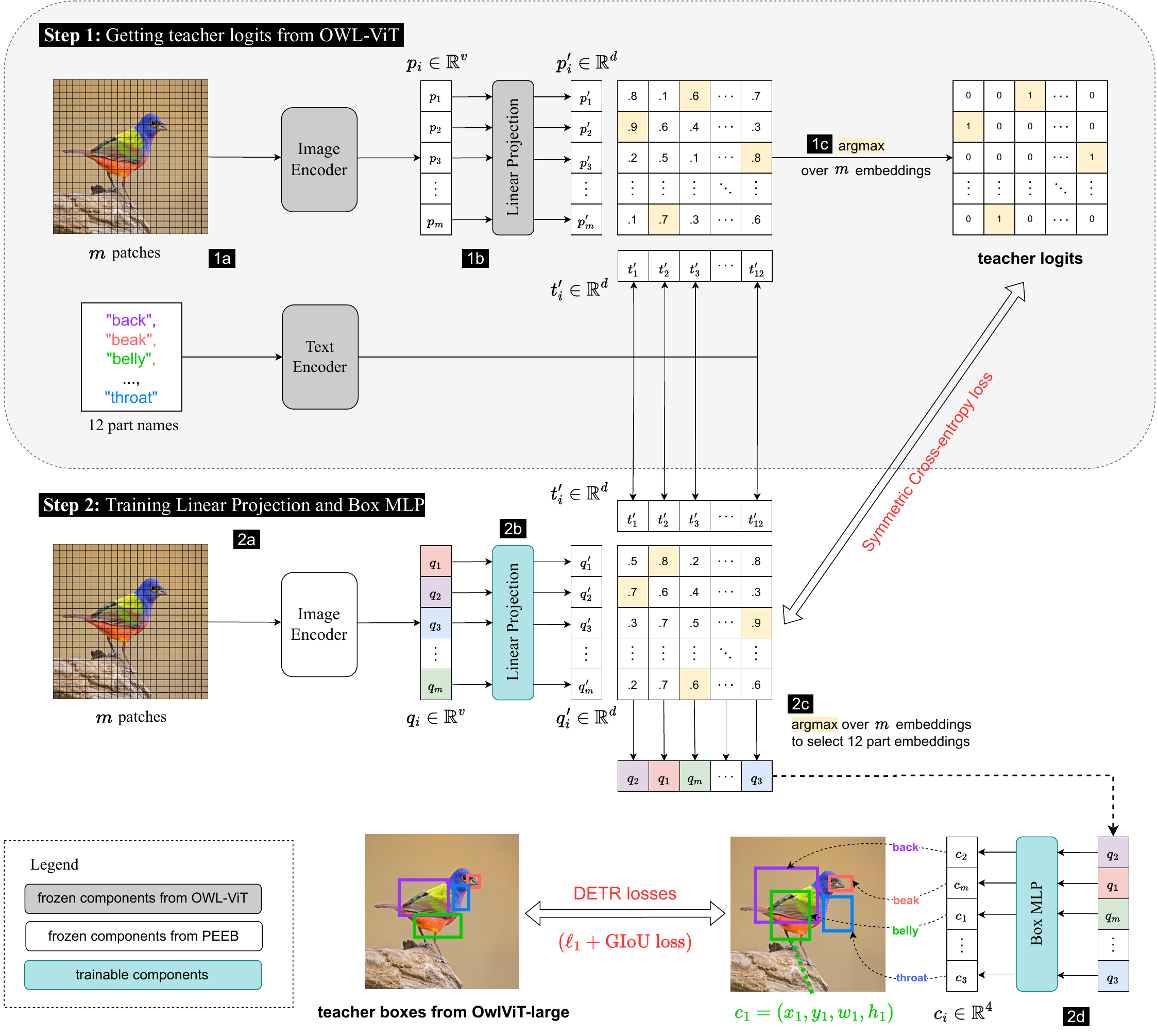}
    \caption{In pre-training stage 2, the goal is to eliminate the teacher model to obtain a standalone classifier. Therefore, the targeted components are \textcolor{TealBlue}{\text{Linear Projection}} and \textcolor{TealBlue}{\text{Box MLP}}. Since these two components are taking care of different functionalities for patch embedding selection and box prediction, respectively, stage 2 training is a multi-objective training. We employ Symmetric Cross-Entropy loss to learn the patch embedding selection and DETR losses to refine the box predictions. In \textbf{Step 1}: (1a), We first encode the 12 part names and the image with \textit{Text Encoder} and \textit{Image Encoder} to obtain the text embedding $\vt'_i$ and patch embedding $\vp_i$. (1b) Then the patch embeddings $\vp$ is projected by \text{Linear Projection} to obtain $\vp'$. (1c) We then compute dot product between $\vp'$  and $\vt'$ and one-hot encode the matrix via the dimension of $\vp'$ to obtain the ``teacher logits''. In \textbf{Step 2}: (2a), We encoder the image with \textit{Image Encoder} to obtain patch embedding $\vq_i$. (2b) The patch embeddings are then being projected by \textcolor{TealBlue}{\text{Linear Projection}} to derive $\vq'$. (2c), We compute the dot product between projected patch embeddings $\vq'$ and part name embeddings $\vt'$ to obtain the similarity matrix. Then, we employ Symmetric Cross-Entropy loss between the similarity matrix and the ``teacher logits'' derived in (1c). (2d), Meanwhile, we select the 12 part embeddings by taking argmax over $\vq'$. Then, the selected part embeddings are forwarded to \textcolor{TealBlue}{\text{Box MLP}} to predict the coordinates of each part. We compute the DETR losses for the predicted coordinates and update the model.}
    \label{fig:pretraining_stage2}
\end{figure*}

As discussed in \cref{sec:training_strategy}, we have three objectives during the \textbf{Pre-training} phase: 

\begin{enumerate}
    \item Pre-training \textbf{Stage 1:} (\cref{fig:pretraining_stage1}) During the pre-training stage one, we contrastively pre-train the model to maximize the similarity between related part-descriptor pairs while minimizing the unrelated pairs using \textit{symmetric cross-entropy (SCE)} loss \citep{radford2021learning}.

    \item Pre-training \textbf{Stage 2:} (\cref{fig:pretraining_stage2}) We try to remove the dependence on the \owlvitLarge teacher model by training PEEB to mimic \owlvitLarge's box predictions using the SCE loss.

    \item Pre-training \textbf{Stage 2:} (\cref{fig:pretraining_stage2}) We simultaneously train PEEB to improve box prediction with DERT losses \citep{zheng2020end}. 
\end{enumerate}

During the \textbf{Finetuning} phase where we finetune on a downstream task (e.g. Dogs-120 or CUB-200), we also employ the same three losses.
However, we change the first loss from SCE into CE since on the downstream classification task, the classifier is tasked with selecting one class that matches the single input image from a set of classes.

\subsubsection{Pre-training stage one: Symmetric cross-entropy loss for contrastive pre-training}

We first define the embeddings derived from the image and text encoders:
\begin{equation}
\label{eq:image_encoder}
I_f' = \text{image\_encoder}(I)
\end{equation}
where $I$ is the input image, and $I_f' \in \mathbb{R}^{n \times d_i}$ is output image embeddings. Here, $d_i$ is the feature dimension of the image encoder. The text embedding $T_f$ is given by 
\begin{equation}
T_f = \text{text\_encoder}(T)
\end{equation}
where $T$ represents the tesxt input, and $T_f \in \mathbb{R}^{m \times d_t}$. In this case, $d_t$ is the feature dimension of the text encoder.
The image embedding $I_f'$ is then transformed by \textit{Part MLP} layer (\cref{fig:xclip_overview}, 1b) to align its dimensions with the text embedding. This transformation is denoted as 
\begin{equation}
I_f = \text{Part MLP}(I_f')
\end{equation}
where $I_f \in \mathbb{R}^{n \times d_t}$.
The similarity matrix $S$ between the image and text embeddings is computed as the dot product of $I_f$ and the transpose of $T_f$, expressed as 
\begin{equation}
S = I_f \cdot T_f^\top
\end{equation}
where $S \in \mathbb{R}^{n \times m}$. The image logits ($S^i$) and text logits ($S^t$) are then defined as
\begin{equation}
S^i = \softmax(S, \text{axis=0})
\end{equation}
and 
\begin{equation}
S^t = \softmax(S, \text{axis=1})
\end{equation}
Next, we define the symmetric cross-entropy loss for the multi-modal embeddings.
\begin{equation}
\label{eq:loss_sce}
L_{sce} = -\frac{(\sum_iy^i_{i}\log(S^i_i) + \sum_my^t_{i}\log(S^t_m)}{2}
\end{equation}
where $y^i \in \mathbb{R}^n$ is the label for image and $y^t \in \mathbb{R}^m$ is the label for text.

\subsubsection{Pre-training stage 2: Symmetric cross-entropy loss to mimic the teacher \owlvitLarge detector}
To mimic the object detection capability of the \owlvitLarge teacher, we train PEEB to mimic the image-text similarity prediction between image embedding and textual part-name embeddings (as shown in \cref{fig:xclip_overview}, 1c). We first binary the teacher logits and consider it as the ground truth label. Then, apply the same symmetric cross-entropy loss as described in \cref{eq:loss_sce} with two minor differences: (1) The text input is part names rather than descriptions. (2) The \textit{Part MLP} is replaced by \textit{Linear Projection} (\cref{fig:xclip_overview}, 2c).

\subsubsection{Pre-training stage 2: DETR losses to mimic the teacher \owlvitLarge detector}
DETR losses are designed to optimize the box detection performance. We employ partial losses in our training for box predictions. Specifically, we employ $\ell_1$ corner-to-corner distance loss and GIoU loss. 
For the selected embeddings, we predict the boxes with \textit{Box MLP} (\cref{fig:xclip_overview}, 2b)
\begin{equation}
    B = \textit{Box MLP}(I_f')
\end{equation}
where $I_f'$ is the image selected image embeddings from \cref{eq:image_encoder}, $B \in \mathbb{R}^{n \times 4}$ is the predicted bounding boxes. Let $Y^{GT} \in \mathbb{R}^{n \times 4}$ be the ground truth boxes. The $\ell_1$ corner-to-corner distance loss is defined as
\begin{equation}
    L_{\ell_1} = \sum_i \left\lVert Y^{GT}_i - B_i \right\rVert
\end{equation}
The GIoU loss $L_{GIoU}$ is defined in \Cref{algo:giou}, and the total box loss is defined as
\begin{equation}
    L_{Box} = \frac{L_{\ell_1} + L_{GIoU}}{2}
\end{equation}

\begin{algorithm}
\label{algo:giou}
\caption{Generalized Intersection over Union}
\begin{algorithmic}[1]
\Require Two arbitrary convex shapes: $A, B \subseteq \mathbb{S} \in \mathbb{R}^n$
\Ensure $GIoU$
\State For $A$ and $B$, find the smallest enclosing convex object $C$, where $C \subseteq \mathbb{S} \in \mathbb{R}^n$
\State $IoU = \frac{|A \cap B|}{|A \cup B|}$
\State $GIoU = IoU - \frac{|C\backslash(A \cup B)|}{|C|}$
\end{algorithmic}
\end{algorithm}

\clearpage

\clearpage

\begin{table*}[ht]
\centering
\setlength\tabcolsep{2.7pt}
\caption{Pre-training details of our pre-trained models.}
\label{tab:preretraining_models}
\resizebox{0.85\textwidth}{!}{%
\begin{tabular}{lccccccccc}\toprule
Model & Epoch & \multicolumn{2}{c}{Batch size} & LR & Weight decay & \multicolumn{2}{c}{\# in-batch classes} & Early stop & Training set \\
\cmidrule(l{2pt}r{2pt}){3-4}
\cmidrule(l{2pt}r{2pt}){7-8}
 &  & Train & Val &  &  & Train & Val &  & \\
\midrule
\multicolumn{10}{c}{\cellcolor{gray!25} Pre-training stage 1} \\
\midrule
\xclipLvOne & 32 & 32 & 50 & $2e^{-4}$ & 0.01 & 48 & 50 & 5 & \birdsoupLvOne \\
\midrule
\xclipLvThreeCUB & 32 & 32 & 50 & $2e^{-4}$ & 0.001 & 48 & 50 & 10 & \birdsoupLvThreeCub \\
\midrule
\xclipLvThreeNABirds & 32 & 32 & 50 & $2e^{-4}$ & 0.001 & 48 & 50 & 10 & \birdsoupLvThreeNABirds \\
\midrule
\multicolumn{10}{c}{\cellcolor{gray!25} Pre-training stage 2} \\
\midrule
\xclipLvOne & 32 & 32 & 50 & $2e^{-5}$ & 0.01 & 48 & 50 & 5 & \birdsoupLvOne \\
\midrule
\xclipLvThreeCUB & 32 & 32 & 50 & $2e^{-5}$ & 0.001 & 48 & 50 & 5 & \birdsoupLvThreeCub \\
\midrule
\xclipLvThreeNABirds & 32 & 32 & 50 & $2e^{-5}$ & 0.001 & 48 & 50 & 5 & \birdsoupLvThreeNABirds \\
\bottomrule
\end{tabular}
}
\end{table*}

\begin{table*}[ht]
\centering
\setlength\tabcolsep{3pt}
\caption{Details of our finetuned models.}
\label{tab:fintuned_models}
\resizebox{0.85\textwidth}{!}{%
\begin{tabular}{lccccccl}\toprule
Model & Fine-tune from & Epoch & Batch size & LR & Weight decay & Early stop & Training set \\
\midrule
\midrule
\xclipLvOneCUB & \xclipLvOne & 30 & 32 & $2e^{-5}$ & 0.001 & 5 & CUB \\
\midrule
\xclipLvThreeCUBClore & \xclipLvThreeCUB  & 5 & 32 & $2e^{-5}$ & 0.001 & 5 & CUB ZSL \citeyearpar{akata2015label} \\
\midrule
\xclipLvThreeCUBSCS & \xclipLvThreeCUB & 5 & 32 & $2e^{-5}$ & 0.001 & 5 & CUB-SCS \\
\midrule
\xclipLvThreeCUBSCE & \xclipLvThreeCUB & 5 & 32 & $2e^{-5}$ & 0.001 & 5 & CUB-SCE \\
\midrule
\xclipLvThreeNABirdsSCS & \xclipLvThreeNABirds & 5 & 32 & $2e^{-5}$ & 0.001 & 5 & NABirds-SCS \\
\midrule
\xclipLvThreeNABirdsSCE & \xclipLvThreeNABirds & 5 & 32 & $2e^{-5}$ & 0.001 & 5 & NABirds-SCE \\
\bottomrule
\end{tabular}
}
\end{table*}

\clearpage

\section{Model and dataset notations}
\label{sec:notations}

\subsection{Dataset notations}
\label{sec:training_sets}
Following the conventional setup of \zsl, we execute certain exclusions to make sure none of the test classes or descriptors are exposed during pre-training. That is, \birdsoupLvThreeCub and \birdsoupLvThreeNABirds exclude all \cub and \nabirds classes, respectively. For \gzsl, we exclude all test sets in \cub, \nabirds, and \inat, denoted as \birdsoupLvOne.
We provide detailed statistics for the three pre-training sets in \Cref{tab:data_splits}. 

\begin{table}[ht]
\centering

\caption{Three pre-training splits for \peeb.}
\label{tab:data_splits}
\resizebox{0.5\textwidth}{!}{%
\begin{tabular}{lcccc}\toprule
Training set & \multicolumn{2}{c}{Number of images} & \multicolumn{2}{c}{Number of classes} \\
\cmidrule(l{2pt}r{2pt}){2-3}
\cmidrule(l{2pt}r{2pt}){4-5}
& Train & Val & Train & Val\\
\midrule
\birdsoupLvOne & 234,693 & 29,234 & 10,740 & 9,746 \\
\midrule
\birdsoupLvThreeCub & 244,182 & 28,824 & 10,602 & 9,608\\
\midrule
\birdsoupLvThreeNABirds & 216,588 & 27,996 & 10,326 & 9,332\\
\bottomrule
\end{tabular}
}
\end{table}

\subsection{Model notations}
We adopt a strategy based on the datasets excluded during training to simplify our model naming convention. Specifically:
\begin{itemize}
    \item \xclipLvOne is pre-trained model using \birdsoupLvOne datset.
    \item \xclipLvThreeCUB is pre-trained model using the \birdsoupLvThreeCub dataset.
    \item \xclipLvThreeNABirds is pre-trained model using the \birdsoupLvThreeNABirds dataset.
\end{itemize}
We named finetuned models after the pre-trained model and the finetuned training set. For example, \xclipLvOneCUB is finetuned from \xclipLvOne, on \cub training set.



\section{Generating part-based descriptors}
\label{sec:descriptors}

CUB annotations initially comprise 15 bird parts.
However, distinctions between the left and right part are not essential to our method, we merge them into a single part (\ie, ``left-wing'' and ``right-wing'' are merged into ``wings'')
Hence, we distilled the original setup into 12 definitive parts: \textit{back, beak, belly, breast, crown, forehead, eyes, legs, wings, nape, tail, throat}.
To compile visual part-based descriptors for all bird species within \birdsoup, we prompted \chatgpt \citep{openai2023gpt4} with the following input template:

\begin{footnotesize}
\begin{prompt}
A bird has 12 parts: back, beak, belly, breast, crown, forehead, eyes, legs, wings, nape, tail and throat. Visually describe all parts of \{class name\} bird in a short phrase in bullet points using the format `part: short phrase'
\end{prompt}    
\end{footnotesize}

Where \texttt{\footnotesize\{class name\}} is substituted for a given bird name (e.g., \class{Painted Bunting}). 

The output is a set of twelve descriptors corresponding to twelve parts of the query species. \eg The response for \texttt{\footnotesize{Cardinal}} is: \\

\begin{lstlisting}[style=nohighlight]
Cardinal: {
    back: vibrant red feathers,
    beak: stout, conical, and orange,
    belly: light red to grayish-white,
    breast: bright red plumage,
    crown: distinctive red crest,
    forehead: vibrant red feathers,
    eyes: small, black, and alert,
    legs: slender, grayish-brown,
    wings: red with black and white accents,
    nape: red feather transition to grayish-white,
    tail: long, red, and wedge-shaped,
    throat: bright red with sharp delineation from white belly
}
\end{lstlisting}

\section{Datasets}
\subsection{\bird \birdsoup}
\label{sec:birdsoup}
We provide a brief statistic of \birdsoup in \cref{tab:birdsoup}. 
\birdsoup is a diverse and long-tailed bird-image dataset. 
The descriptors generated by GPT-4 are in English and only describe the visual features of the corresponding class. We propose \birdsoup for academic research only.

\begin{table}[ht]
\centering
\caption{Number of images and species of different bird datasets. Our proposed dataset Bird-11K includes almost all avians on Earth.} 
\label{tab:birdsoup}
\resizebox{0.8\textwidth}{!}{
\begin{tabular}{lrr}
\toprule
\textbf{Dataset} & \# of \textbf{Images} & \# of \textbf{Species}\\
\midrule
CUB-200-2011 \citep{wah2011caltech} & 12,000 & 200 \\ 
Indian Birds \citep{kaggle_ebird_dataset} & 37,000 & 25 \\
NABirds v1 \citep{van2015building} & 48,000 & 400 \\ 
Birdsnap v7 \citep{berg2014birdsnap} & 49,829 & 500 \\ 
\href{https://github.com/visipedia/inat_comp/tree/master/2021}{iNaturalist 2021-birds} \citep{van2021benchmarking} & 74,300 & 1,320 \\
ImageNet-birds \citep{deng2009imagenet} & 76,700 & 59  \\ 
BIRDS 525 \citep{birds525} & 89,885 & 525 \\
Macaulay Library at the Cornell Lab of Ornithology\tnote{*} & 55,283 & 10,534 \\ \midrule
\birdsoup (Raw Data) & 440,934 & 11,097 \\ 
\textbf{\birdsoup (pre-training set)} & 294,528 & 10,811 \\ \bottomrule
\end{tabular}
}
\end{table}

\paragraph{Data splits}
We provide data splits and metadata, \eg, file names, image size, and bounding boxes, along with the instruction of \birdsoup construction in our repository. Note that the \birdsoup dataset is for pre-training purposes; it is important to execute exclusion based on the test set.

\paragraph{License and terms} 
\begin{itemize}
    \item CUB \citep{wah2011caltech}: The dataset can be freely used for academic and research purposes; commercial use is restricted.
    \item Indian Birds \citep{kaggle_ebird_dataset}: CC0: Public Domain.
    \item NABirds-v1 \citep{van2015building}: For non-commercial research purposes, other use is restricted \footnote{See \href{https://dl.allaboutbirds.org/merlin---computer-vision--terms-of-use?submissionGuid=4edd06f5-55b9-4050-a935-6054737e4a9f}{Terms of Use}} here for detail: . 
    \item Birdsnap-v7 \citep{berg2014birdsnap}: The dataset creator provides no specific license or terms of use. We only use this dataset for academic research until more specific details can be obtained.
    \item iNaturalist 2021-birds \citep{van2021benchmarking}: CC0: Public Domain. We use the \texttt{train\_mini} subset on \href{https://github.com/visipedia/inat_comp/tree/master/2021}{Github}, which has 1,486 classes. After filtering out images (as described in \cref{sec:bird11_construction}), we end up with 1,320 classes and 74,300 images for including in Bird-11K.
    \item ImageNet-birds \citep{deng2009imagenet}: BSD-3-Clause license.
    \item BIRDS 525  \citep{birds525}: CC0: Public Domain
    \item Cornell eBird: We used the following 55,384 recordings from the Macaulay Library at the Cornell Lab of Ornithology. The data is for academic and research purposes only, not publicly accessible unless requested. (Please refer to our Supplementary Material for the full list):

\tiny ML187387391, ML187387411, ML187387421, ML187387431, ML262407521, ML262407481, ML262407531, ML262407491, ML262407511, ML257194111
\tiny ML257194071, ML257194081, ML257194061, ML495670791, ML495670781, ML495670801, ML495670771, ML183436431, ML183436451, ML183436441
\tiny ML183436411, ML183436421, ML256545901, ML256545891, ML256545841, ML256545851, ML256545831, ML169637941, ML238083081, ML169637881
\tiny ML169637911, ML238083111, ML238083051, ML169637971, ML299670841, ML64989231, ML299670831, ML64989241, ML299670791, ML64989251
\tiny ML246866001, ML246865941, ML246866011, ML246865961, ML246865971, ML333411961, ML240835531, ML240835541, ML240835701, ML240835591
\tiny ML245260391, ML245260341, ML245260371, ML245260411, ML245260421, ML245260431, ML245260441, ML240866351, ML240866331, ML240866321
\tiny ML240866341, ML240866371, ML248318661, ML248318571, ML248318591, ML248318581, ML248318631, ML245204281, ML245204311, ML245204371
\tiny ML245204381, ML245204291, ML245603571, ML245603521, ML245603511, ML245603491, ML245603501, ML245603601, ML245257771, ML245257651
\tiny ML245257631, ML245257661, ML245257761, ML247221051, ML247221061, ML247221071, ML247221081, ML240365811, ML240365751, ML240365781
\tiny ML240365761, ML300579541, ML247298551, ML247298541, ML247298561, ML247298611, ML247298571, ML247298591, ML247298601, ML247298631...
    
\end{itemize}

\subsection{\dog Dog-140}
\label{sec:dogsoup}
\begin{figure}[!hbt]
    \centering
    \includegraphics[width=0.8\linewidth]{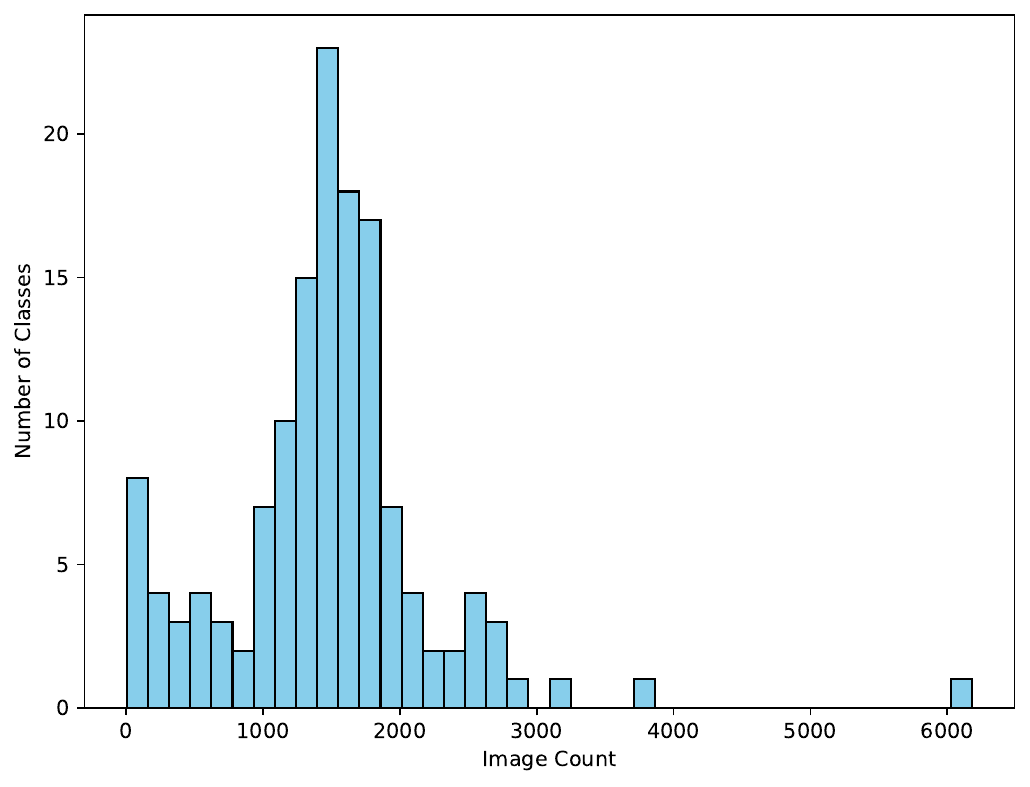}
    \caption{The class distribution of Dog-140 dataset. The histogram indicates that most classes in Dog-140 have around 1,000 to 2,000 images.}
    \label{fig:dogsoup_v1_distribution}
\end{figure}

To pre-train \peeb on dogs, we construct Dog-140 by combining dog images from ImageNet-21K and Stanford Dogs-120.
Specifically, we selected 189 dog classes from ImageNet-21K, and based on Fédération Cynologique Internationale (FCI) \citep{FCINomenclature2023}, we merged them with 120 classes from Stanford Dogs, ending up with 140 classes. After merging, Dog-140 has 206,076 images in total. We provide a class distribution analysis in \Cref{fig:dogsoup_v1_distribution}, where we can find that Dog-140 is roughly class-balanced.

\paragraph{Data splits}
Similar to \birdsoup, we provide data splits and metadata, \eg, file names, image size, and bounding boxes, along with the instruction of Dog-140 construction in our repository.

\paragraph{License and terms} 
\begin{itemize}
    \item Stanford Dogs \citep{khosla2011novel}: The dataset was constructed using images and annotations from ImageNet. Therefore, all the images (including those presented in the paper) follow the ImageNet license.
    \item ImageNet-21K \citep{deng2009imagenet}: BSD-3-Clause license, non-commercial.
\end{itemize}


\section{Additional results}
\label{apx:ablation_study}

\subsection{\peeb outperforms \mv in CUB and NABirds in ZSL setting}
\label{abl:stress_test}

To rigorously evaluate the \zsl capabilities of our pre-trained models, we introduce a stress test on the \cub and \nabirds datasets. The crux of this test involves excluding all classes from the target dataset (CUB or NABirds) during the pre-training. The exclusion ensures that the model has no prior exposure to these classes. Subsequently, we measure the classification accuracy on the target dataset, comparing our results against benchmarks set by CLIP and \mv in the scientific name test. In this experiment, we consider the scientific name test a \zsl test for CLIP and use them as the baseline because the frequencies of scientific names are much lower than common ones. 

\paragraph{Experiment}
To conduct this test, we pre-train our model on \birdsoupLvThreeCub and \birdsoupLvThreeNABirds, which deliberately exclude images bearing the same class label as the target dataset. Specifically, we test on our pre-train model \xclipLvThreeCUB and \xclipLvThreeNABirds (see \Cref{tab:preretraining_models} for details), respectively.

\paragraph{Results}
The primary objective is to ascertain the superiority of our pre-trained model, \peeb, against benchmarks like CLIP and \mv. For CUB, our method reported a classification accuracy of 17.9\%, contrasting the 5.95\% and 7.66\% achieved by CLIP and \mv, respectively, as shown in \cref{tab:stress_test}. 
The \peeb score, which is substantially higher (\increasenoparent{10}) than \mv, highlights the advantages of our part-based classification.
On \nabirds, our method surpasses CLIP and \mv by \increasenoparent{1} point. The performance disparity between \cub and \nabirds can be attributed to two factors: The elevated complexity of the task (555-way classification for \nabirds versus 200-way for \cub) and the marked reduction in training data. An auxiliary observation, detailed in \Cref{abl:effect_of_classes_and_images}, indicates that our pre-trained model necessitates at least 250k images to achieve admirable classification accuracy on CUB, but we only have 210k images training images in \birdsoupLvThreeNABirds (the variants of Bird-11K with classes excluded for \zsl testing are described in \Cref{tab:data_splits}).

\begin{table}[hbt]
\centering
\caption{Stress test results on \cub and \nabirds datasets. Despite the \zsl challenge, our method consistently surpasses CLIP and \mv. This underscores the robust generalization of our approach, which leverages descriptors for classification.}
\label{tab:stress_test}
\resizebox{0.35\textwidth}{!}{%
\begin{tabular}{lccc}\toprule
Method & CLIP & \mv & \peeb (ours) \\
\midrule
\midrule
\cub & 5.95 & 7.66 & \textbf{17.90}\\
\midrule
\nabirds & 4.73 & 6.27 & \textbf{7.47} \\
\bottomrule
\end{tabular}
}
\end{table}

\subsection{Performance measurement on different noisy levels}
\label{abl:inat_noise_test}

In our evaluations, as indicated in \cref{tab:clip_like_zeroshot}, we discerned a marked performance disparity between the \inat dataset and others. Probing this further, we identified image noise as a principal contributor to these discrepancies.

\paragraph{Experiment}
A qualitative assessment of the \inat test images revealed a significantly higher noise level than \cub or \nabirds. To systematically study this, we utilize the object detector \owlvitLarge to measure the size of the bird within the images. We formulated two filtered test sets based on the detector's output, categorizing them by the bird's size, specifically, the detected bounding box. Images were filtered out if the bird's size did not exceed predetermined thresholds (areas of $100^2$ or $200^2$ pixels). Larger birds naturally reduced other content by occupying more image space, thus serving as a proxy for reduced noise. All three test sets, including the original, were evaluated using our pre-trained model \xclipLvOne.

\paragraph{Results}
The results presented in \cref{tab:inat_noise_test} reveal a clear trend: as the image noise level decreases, the classification accuracy consistently improves, with gains ranging from (\increasenoparent{6} to \increasenoparent{17}) points across the various methods. Notably, cleaner images consistently yield better results. At each noise level, our method outperforms the alternatives. While our method exhibits an impressive (\increasenoparent{17} points) accuracy boost on the cleanest test set, this substantial gain also indicates that our model is sensitive to image noise.

\begin{table}[hbt]
\centering
\caption{The table showcases the classification accuracies on iNaturalist as we vary the noise levels. The data underscores that the performance disparity on iNaturalist is predominantly due to image noise. While all methods improve with cleaner images, our model exhibits the most substantial gains, particularly in the least noisy sets.}
\label{tab:inat_noise_test}
\resizebox{0.4\textwidth}{!}{%
\begin{tabular}{lccc}\toprule
Splits & CLIP & \mv & \peeb (ours) \\
\midrule
\midrule
Original & 16.36 & 17.57 & \textbf{25.74}\\
\midrule
$> 100^2$ pixels & 20.18 & 21.66 & \textbf{35.32} \\
\midrule
$> 200^2$ pixels & 22.88 & 24.90 & \textbf{42.55} \\
\bottomrule
\end{tabular}
}
\end{table}

\subsection{Number of training images is the most critical factor towards classification accuracy}
\label{abl:effect_of_classes_and_images}

\begin{figure}[!hbt]
    \centering
    \begin{subfigure}[b]{0.8\linewidth}
        \includegraphics[width=\linewidth]{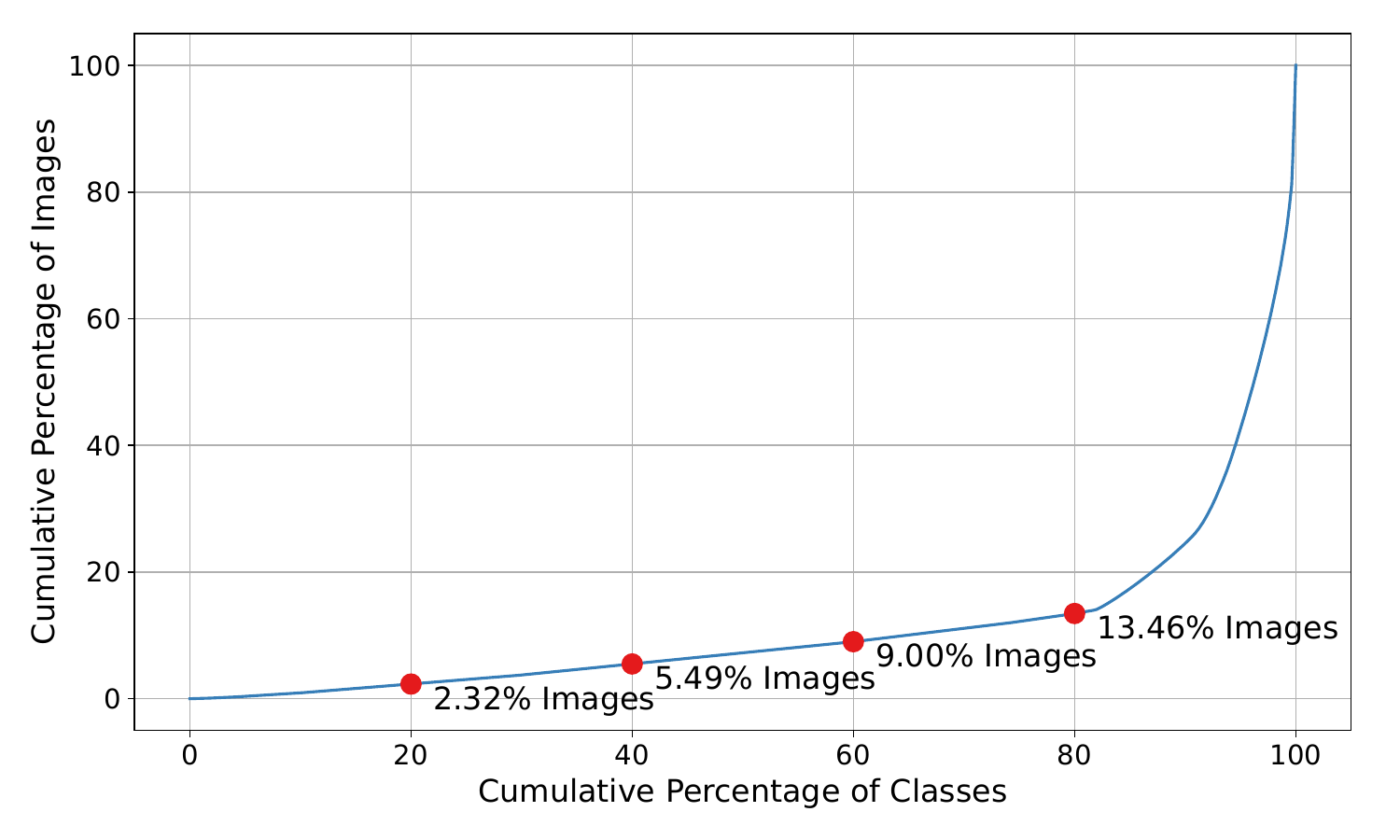}
        \caption{The Cumulative Distribution Function (CDF) plot for the \birdsoup dataset.}
        \label{fig:abl_bs_stats}
    \end{subfigure}
    \begin{subfigure}[b]{0.8\linewidth}
        \includegraphics[width=\linewidth]{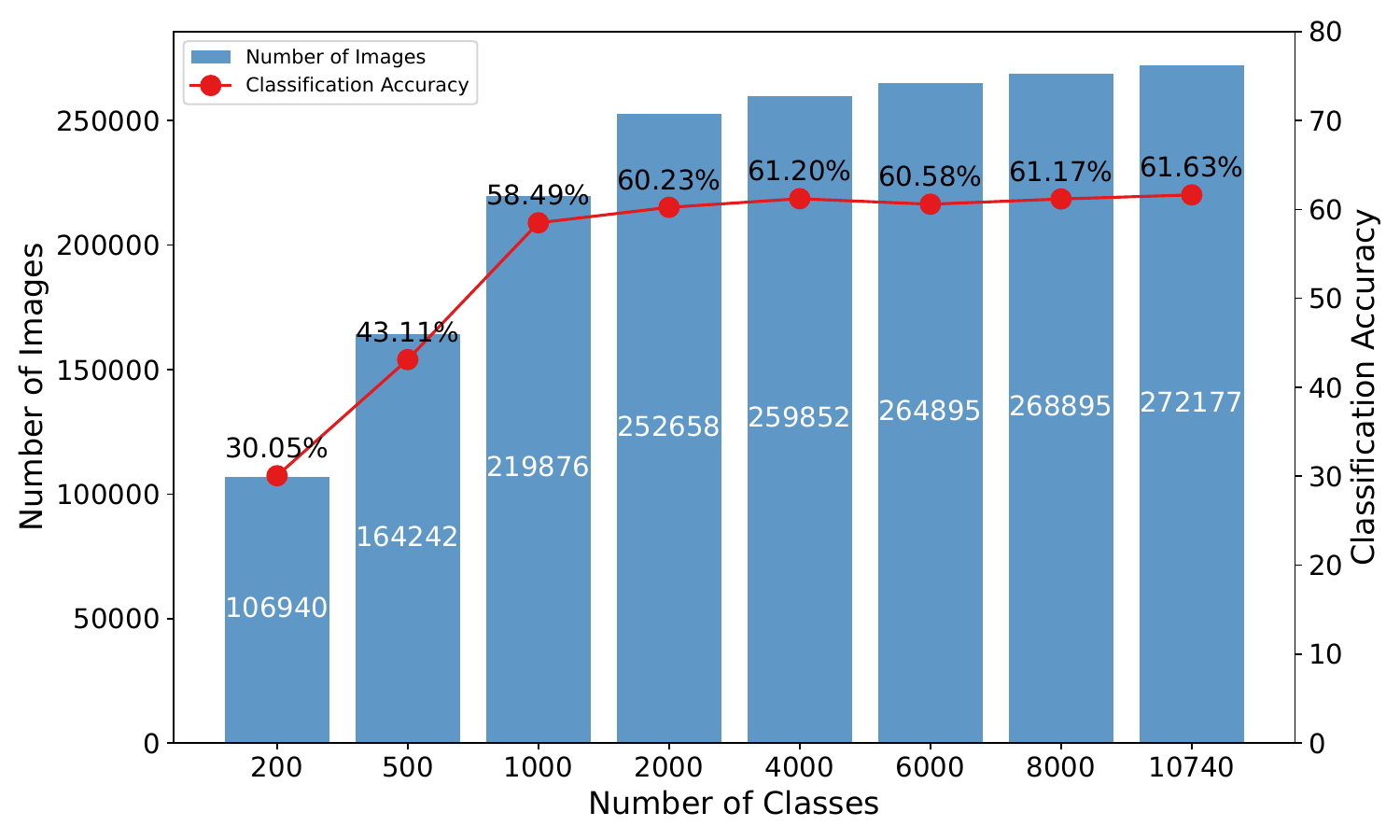}
        \caption{Correlation between the number of training images/classes and accuracy.}
        \label{fig:abl_effect_of_classes_and_images}
    \end{subfigure}
    \caption{The CDF plot (a), underscores significant imbalance of the \birdsoup dataset. While the dataset has abundant long-tailed classes, e.g., a striking 80\% of the classes contribute to only 13.46\% of the entire image count. The plot (b) showcases the correlation between the number of training images/classes and the resulting classification accuracy. As the image count grows, there is a noticeable surge in accuracy, which nearly stabilizes upon surpassing 250K images. Additionally, a significant amount of long-tailed data contributes to a \increasenoparent{1.5} points boost in accuracy.}
    \label{fig:combined_figure}
\end{figure}

\birdsoup, as shown in \cref{fig:abl_bs_stats}, is a highly imbalanced dataset characterized by a large amount of long-tailed classes. We conduct a comprehensive study to discern how variations in the number of classes and images affect the classification accuracy of our pre-trained models. Predictably, the volume of training images occurred as the most influential factor. However, a noteworthy observation was that the abundance of long-tailed data enhanced the model's accuracy by approximately \increasenoparent{1.5} points.

\paragraph{Experiment}
We curated eight training sets based on varying class counts: 200, 500, 1,000, 2,000, 4,000, 6,000, 8,000, and 10,740. For each set, we maximized the number of training images. It is important to note that a set with a lesser class count is inherently a subset of one with a higher count. For instance, the 500-class set is a subset of the 2,000-class set. For each split, we apply the same training strategy as in \Cref{sec:pretraining}, and choose the checkpoint with the best validation accuracy.  We consider the \cub test set as a generic testing benchmark for all variants.

\paragraph{Results}
As illustrated in Figure \cref{fig:abl_effect_of_classes_and_images}, there is a pronounced correlation between the increase in the number of images and the corresponding surge in accuracy. For instance, an increment from 106K to 164K images led to a rise in classification accuracy from 30.05\% to 43.11\%. The accuracy appears to stabilize around 60\% when the image count approaches 250K. This trend strongly suggests that the volume of training images is the most critical factor for the pre-trained model. We believe that the accuracy of the pre-trained model could be further enhanced if enough data is provided. Interestingly, a substantial amount of long-tailed data bolsters the model's performance, evident from \increasenoparent{1.5} points accuracy improvement when comparing models trained on 2,000 classes to those on 10,740 classes. Note that the additional classes in the latter set averaged merely 2.2 images per class.


\FloatBarrier
\FloatBarrier

\subsection{Ablation study on the influence of parts utilized}
\label{abl:effect_on_number_of_parts}

In this ablation study, we aimed to measure the impact of varying the number of distinct ``parts'' (back, beak, belly, breast, crown, forehead, eyes, legs, wings, nape, tail, and throat) used in our model. We experiment with a range from a single part to all 12 identifiable parts. Interestingly, even with a solitary part, the model could make correct predictions, though there was an evident decline in performance, approximately \decreasenoparent{20} points.

\paragraph{Experiment}
Our testing ground is the pre-trained model \xclipLvOne, evaluated against the \cub test set. We assessed the model's prowess utilizing various subsets of parts: 1, 3, 5, 8, and all 12. These subsets were derived based on the frequency of visibility of the parts within the \cub dataset, enabling us to compare the model's performance when relying on the most frequently visible parts versus the least. For comparison, we also conduct a similar experiment on \mv, where we only use 1, 3, 5, 8, and 12 descriptors (if possible).

\paragraph{Results}
Relying solely on the most frequent part led to a decline in classification accuracy by around \decreasenoparent{20} points, registering at 45.44\% (\cref{tab:effects_on_num_parts}). In contrast, utilizing the least frequent part resulted in a sharper drop of around \decreasenoparent{27}, with an accuracy of 37.02\%. As the model was furnished with increasing parts, its accuracy improved incrementally. The data underscores that optimal performance, an accuracy of 64.33\%, is attained when all 12 parts are included. For \mv, the accuracy keeps increasing homogeneously from 5 to 12 descriptors, hinting that accuracy may increase further by increasing the number of descriptors.

\begin{table*}[hbt]
\centering 
\caption{Classification accuracy on the \cub test set that uses a different number of parts. Performance dips significantly with just one part, especially for the least visible ones. Maximum accuracy is reached with all 12 parts. The last row of the table also shows the accuracy of \cite{menon2023visual} method which employs a different number of parts. It is evident that their method is insensitive to the number of parts used, which may not reflect a realistic scenario.}
\label{tab:effects_on_num_parts}
\resizebox{0.7\textwidth}{!}{%
\begin{tabular}{lccccc}\toprule
Number of Parts (descriptors) & 1 & 3 & 5 & 8 & 12 \\
\midrule
\midrule
Accuracy (most frequent parts) & 45.44 & 56.48 & 59.89 & 61.32 & \textbf{64.33} \\
\midrule
Accuracy (least frequent parts) & 37.02 & 55.51 & 60.04 & 61.13 & \textbf{64.33} \\
\midrule
Accuracy of \cite{menon2023visual} & 51.93 & 52.87 & 52.83 & 53.33 & 53.92 \\
\bottomrule
\end{tabular}
}
\end{table*}

\subsection{Training is essential for \peeb's classification efficacy}
\label{sec:xclip_no_training}
In this ablation study, we highlight the pivotal role of training in the performance of \peeb on bird classification tasks. We demonstrate that without adequate tuning, the results are indistinguishable from random chance.

\paragraph{Experiment}
We conduct the experiment based on \owlvitBase. We retain all components as illustrated in \Cref{fig:xclip_overview}, with one exception: we substitute the Part MLP with the MLP layer present in the box prediction head of \owlvit because the proposed layers require training. The MLP layers in the box prediction head project the part embeddings to match the dimensionality of the text embeddings. Our focus is on assessing the classification accuracy of the untuned \peeb on two datasets: \cub and \nabirds.

\paragraph{Results}
\Cref{tab:xclip_no_training} reveals the outcomes of our experiment. Without training, \peeb yields classification accuracies of 0.55\% for \cub and 0.31\% for \nabirds, both of which are proximate to random chance (0.5\% for \cub and 0.1\% for \nabirds). However, with training, the model's performance dramatically transforms: 64.33\% for \cub (an increase of \increasenoparent{63.78} points) and 69.03\% for \nabirds (a leap of \increasenoparent{68.72} points) for \xclipLvOne. These pronounced disparities underscore the vital role of training in \peeb.

\begin{table}[hbt]
\centering 
\caption{Impact of Training on Classification Accuracies: Untuned \peeb yields 0.55\% on \cub and 0.31\% on \nabirds, almost mirroring random chance. With training (\xclipLvOne), accuracy surges by \increasenoparent{63.78} points on \cub and \increasenoparent{68.72} points on \nabirds.}
\label{tab:xclip_no_training}
\resizebox{0.39\textwidth}{!}{%
\begin{tabular}{lcc}\toprule
 & \cub & \nabirds  \\
\midrule
\midrule
\peeb (no training) & 0.55 & 0.31 \\
\midrule
\xclipLvOne pre-trained & 64.33 & 69.03 \\
\midrule
\xclipLvOneCUB finetuned & 86.73 & - \\
\bottomrule
\end{tabular}
}
\end{table}

\subsection{Failure analysis}

Since \peeb has two branches, box detection, and descriptor matching, we would like to find out, in the failure case, what is the main cause.
i.e., is it because of the mismatch in the descriptor to the part embeddings? Or is it because the box detection is wrong? From our ablation study, it turns out that most errors come from the descriptor-part matching.

\paragraph{Experiment}
We conduct the experiment with \xclipLvOne on \cub test set. Specifically, we measure the box detection accuracy based on the key point annotation in \cub dataset, i.e., We consider the box prediction as \textbf{correct} if the prediction includes the human-annotated key point. We report the box prediction error rate (in \%) based on parts. 

\paragraph{Results}
As shown in \cref{tab:box_prediction_error}, the average error rate difference between success and failure cases is merely 0.38. That is, in terms of box prediction, the accuracy is almost the same, disregarding the correctness of bird identification. It indicates that the prediction error is predominantly due to the mismatch between descriptors and part embeddings. We also noted that some parts, like \class{Nape} and \class{Throat}, have a very high average error rate, which may greatly increase the matching difficulties between descriptors and part embeddings.

\begin{table*}[hbt]
\centering
\caption{Error rate of Box Prediction in Failure and Success Cases. We report the box prediction error rate, depending on whether the prediction box includes ground truth key points. No major difference is found between them, which means the failure is largely due to the part-descriptor mismatch.}
\label{tab:box_prediction_error}
\resizebox{0.99\textwidth}{!}{
\begin{tabular}{lccccccccccccc}\toprule
\textbf{Body Part} & \textbf{Average} & \textbf{Back} & \textbf{Beak} & \textbf{Belly} & \textbf{Breast} & \textbf{Crown} & \textbf{Forehead} & \textbf{Eyes} & \textbf{Legs} & \textbf{Wings} & \textbf{Nape} & \textbf{Tail} & \textbf{Throat} \\
\midrule
Failure Cases & 16.52 & 23.38 & 3.28 & 8.06 & 15.96 & 7.41 & 24.72 & 7.29 & 5.63 & 3.36 & 64.79 & 7.25 & 27.07 \\
Success Cases & 16.14 & 23.03 & 2.96 & 7.44 & 18.64 & 7.13 & 21.53 & 3.93 & 6.85 & 2.68 & 68.66 & 6.40 & 24.38 \\
Difference    & \textbf{0.38}  & 0.35  & 0.33  & 0.62  & -2.68 & 0.28 & 3.19  & 3.36  & -1.22 & 0.68  & -3.87 & 0.85  & 2.68  \\
\bottomrule
\end{tabular}
}
\end{table*}

\subsection{Evaluation of predicted boxes from \peeb}
\label{sec:box_evaluation}

Our proposed method primarily aims to facilitate part-based classification. 
While the core objective is not object detection, retaining the box prediction component is paramount for ensuring model explainability. 
This section delves into an evaluation of the box prediction performance of our method against the \owlvitBase model. 

\paragraph{Experiment}
Given our focus on part-based classification, we aimed to ascertain the quality of our model's box predictions. 
To this end, we employed two metrics: mean Intersection over Union (IoU) and precision based on key points.
We opted for mean IoU over the conventional mAP because: (1) Ground-truth boxes for bird parts are absent, and (2) our model is constrained to predict a single box per part, ensuring a recall of one. 
Thus, we treat \owlvitLarge's boxes as the ground truth and evaluate the box overlap through mean IoU. Furthermore, leveraging human-annotated key points for bird parts, we measure the precision of predicted boxes by determining if they contain the corresponding key points. 
We evaluate our finetuned models on their corresponding test sets. 
For instance, \xclipLvThreeCUBClore, finetuned based on the CUB split \citep{akata2015label}, is evaluated on the \cub test set.

\paragraph{Results}
Our evaluation, as presented in \cref{tab:model_evaluation}, shows that \peeb's box predictions do not match those of \owlvitBase. 
Specifically, on average, there is a \decreasenoparent{5} to \decreasenoparent{10} points reduction in mean IoU for \cub and \nabirds datasets, respectively. The disparity is less distinct when examining precision based on human-annotated key points; our method records about \decreasenoparent{0.14} points lower precision for \cub and \decreasenoparent{3.17} points for \nabirds compared to those for \owlvitBase. 
These observations reinforce that while \peeb's box predictions might not rival these dedicated object detection models, they consistently highlight the same parts identified by such models as shown in \cref{fig:box_comparison}. 
It is important to note that our approach utilized the same visual embeddings for both classification and box prediction tasks. 
This alignment emphasizes the part-based nature of our model's predictions.

\begin{table}[htb]
\centering
\caption{Model evaluation on \cub and \nabirds test sets. 
We evaluate the predicted boxes on two \textit{ground-truth} sets; (1) predicted boxes from \owlvitLarge as ground-truths, and (2) \owlvitLarge's boxes that include the human-annotated key points. 
Our method has slightly lower performance in terms of mean IoU but comparable precision.}
\label{tab:model_evaluation}
\resizebox{0.6\textwidth}{!}{
\begin{tabular}{llrrc}
\toprule
& \multirow{2}{*}{Models} & \multicolumn{2}{c}{Mean IoU} &  \\
\cmidrule(l{2pt}r{2pt}){3-4}
& & (1) All & (2) w/ Keypoints & Precision\\
\midrule
\cub & \owlvitLarge & \textbf{100.00} & \textbf{100.00} & \textbf{83.83} \\
 & \owlvitBase & 44.41 & 49.65 & 83.53 \\
 & \peeb (Average) & 35.98 & 40.14 & 83.39 \\
\cmidrule{3-5}
 & \xclipLvOneCUB & 37.45 & 41.79 & 81.55 \\
 & \xclipLvThreeCUBClore & 35.11 & 39.14 & 82.72 \\
 & \xclipLvThreeCUBSCS & 35.77 & 39.96 & 84.89 \\
 & \xclipLvThreeCUBSCE & 35.58 & 39.67 & 84.38 \\

\midrule
\nabirds & \owlvitLarge & \textbf{100.00} & \textbf{100.00} & \textbf{85.01} \\
 & \owlvitBase & 40.14 & 47.63 & 83.89 \\
 & \peeb (Average) & 36.47 & 42.01 & 80.72 \\
\cmidrule{3-5}
 & \xclipLvThreeNABirdsSCS & 36.45 & 42.03 & 80.09 \\
 & \xclipLvThreeNABirdsSCE & 36.49 & 41.99 & 81.34 \\
\bottomrule
\end{tabular}
}
\end{table}

\clearpage
\section{Study on GPT-4 generated descriptors}

\subsection{Assessment of the generated part-based descriptors}
\label{sec:descriptors_assessment}
We test GPT-4V on the CUB test set using the generated descriptors of 200 classes to assess their usability.
Specifically, we feed GPT-4V with each test image encoded in the payload and 200 sets of part-based descriptors through a carefully designed prompt (\Cref{tab:gpt4v_prompt}).
GPT-4V is asked to output one of 200 provided class names to compute the classification accuracy.
As a result, GPT-4V achieves 69.4\% accuracy which is slightly higher than \peeb's generalized zero-shot accuracy (64.33\%) and significantly lower than \peeb results after finetuning (86-88\%).

\begin{table}[htb]
\centering
\caption{Prompt for GPT-4V evaluation on \cub where {\{list\_of\_200\_classes\}} is the placeholder for the actual 200 CUB classes while {\{descriptors\}} (see an example in \cref{sec:descriptors}) is the placeholder for the actual descriptors associated with a given bird image from the CUB test set.}
\label{tab:gpt4v_prompt}
\begin{tcolorbox}
\footnotesize
\begin{tabular}{p{\textwidth}}
You are an image classifier which can tell what type of a bird is from the given image and its associated part descriptors describing 12 parts of the bird.
Your answer should be strictly formatted as \{"prediction": "bird\_class"\}.\\
\\
where "bird\_class" is one of the following 200 bird classes:
\{list\_of\_200\_classes\}\\
\\
Given the bird image and the following descriptors:
\{descriptors\}\\
\\
What kind of bird is this? Let's think step by step.
\end{tabular}
\end{tcolorbox}
\end{table}

\subsection{Noise measurement in GPT-4 generated descriptors}
\label{sec:gpt_noise_measure}
In this section, we conduct an empirical analysis to quantify the noise in descriptors generated by GPT-4 for 20 different classes within the CUB dataset. To achieve this, we manually inspect each descriptor and tally the instances where at least one factual error is present. Our findings reveal that every one of the 20 classes contains descriptors with errors, and on average, 45\% of the descriptors necessitate corrections. This substantial noise level underscores the need for further refinement in our work, particularly in text descriptors.

We observe a notably high error rate in descriptors on the \textit{back} and \textit{wings}, with approximately 60\% of these containing inaccurate information (refer to \Cref{tab:gpt_noise_measure}). This could be attributed to the challenges in distinguishing between the \textit{back} and \textit{wings}, given that the \textit{back} is typically positioned behind the \textit{wings}, yet exhibits considerable variability in size and shape. Addressing all descriptor issues by revising all 11,000 fine-grained descriptors would demand a significant investment of time and resources, which is beyond the scope of the current work. As such, we identify this as an area for future research and development, aiming to enhance the quality of the \birdsoup dataset.

\begin{table*}[hbt]
\centering 
\caption{Summary of manual inspection results for 20 classes, highlighting the need for revision in GPT-4 generated descriptors. An average error rate of 45\% indicates substantial room for improvement.}
\label{tab:gpt_noise_measure}
\resizebox{0.99\textwidth}{!}{
\begin{tabular}{lccccccccccccc}\toprule
  & Back & Beak & Belly & Breast & Crown & Forehead & Eyes & Legs & Wings & Nape & Tail & Throat & Average\\
\midrule
\midrule
Error Rate & 60 & 30 & 50 & 40 & 50 & 55 & 50 & 20 & 60 & 50 & 35 & 40 & 45 \\
\bottomrule
\end{tabular}
}
\end{table*}

\subsection{Revising descriptors improves classification accuracy}
\label{sec:revise_description}
As mentioned in the limitation section, the descriptors are generated from GPT-4 and therefore noisy and incorrect. 
Given that \peeb accepts open vocabulary inputs for classification, a natural way to improve classification accuracy is to improve the correctness of the descriptors.

\paragraph{Experiment}
We first collect descriptors of 183 \cub classes from AllAboutBirds. We then prompt GPT-4 to revise our original descriptors by providing the collected descriptor. We revise the descriptors with the following prompt:

\begin{footnotesize}
\begin{prompt}
Given the following descriptors of \{class name\}: \{AllAboutBirds descriptors\}. Can you revise the incorrect items below (if any) of this bird, return them as a Python dictionary, and use the key as the part name for each item? If a part\'s descriptor is not specifically described or cannot be inferred from the definition, use your own knowledge. Otherwise, leave as is. Note: please use a double quotation mark for each item such that it works with JSON format.

\{Original descriptors\}
\end{prompt}    
\end{footnotesize}
Where \texttt{\footnotesize\{class name\}} the placeholder for the class name, \texttt{\footnotesize\{AllAboutBirds descriptors\}} is the description collected from AllAboutBirds, \texttt{\footnotesize\{Original descriptors\}} is the descriptors we used for training.

 Due to the errors in the descriptors we used to train \peeb, simply replacing the descriptors with their revised version does not lead to better performance. Because the incorrect descriptors in training change the meaning of some of the phrases. For example, the belly of \class{Blue bunting} is pure blue, but the descriptors from GPT-4 is \textit{soft, creamy white}. In addition, the GPT-4 uses the exact same descriptor in the belly for other classes, e.g., \class{Blue breasted quail}, which should be cinnamon. \class{Blue Fronted Flycatcher}, which should be yellow. Training the same descriptors with different colors confuses the model, and the model will convey the phrase ``creamy white'' with a different meaning to humans. Therefore, simply changing the descriptors to their' revised version will not work. We empirically inspect the descriptors that \peeb can correctly respond to and replace the class descriptors with the revised version. Specifically, we replace the descriptors of 17 classes in \cub and test the classification accuracy on \xclipLvOne. 

\paragraph{Results}
As shown in \Cref{tab:effects_on_revise_desc}, the overall accuracy increases by \increasenoparent{0.8} points. 

The average improvement of the revised class is around \increasenoparent{10.8}, hitting that if we have correct descriptors of all classes, we may significantly improve the classification accuracy of the pre-trained model. However, correcting all 11k class descriptors is too expensive and out of the scope of this work. We leave it as a further direction of improving the part-based bird classification.

\begin{table}[hbt]
\centering 
\caption{The revised descriptors result in \increasenoparent{0.8} for \xclipLvOne in \cub. In particular, the average improvement among the 17 revised classes is \increasenoparent{10.8}, hinting at the large potential of our proposed model.}
\label{tab:effects_on_revise_desc}
\resizebox{0.5\textwidth}{!}{%
\begin{tabular}{lccc}\toprule
Descriptors & Original & Partially Revised & Avg. Improvement \\
\midrule
\midrule
\xclipLvOne & 64.33 & \textbf{65.14} & 10.80 \\
\bottomrule
\end{tabular}
}
\end{table}


\clearpage
\section{Qualitative Inspections}
\label{sec:qualitative_examples}

\subsection{Visual comparison of predicted boxes} 
\begin{figure*}
    \centering
    \makebox[0.25\textwidth][c]{Original}%
    \makebox[0.25\textwidth][c]{\peeb}%
    \makebox[0.25\textwidth][c]{\owlvitBase}%
    \makebox[0.25\textwidth][c]{\owlvitLarge}
    
    \begin{subfigure}{\textwidth}
        \centering
        \includegraphics[width=1\linewidth]{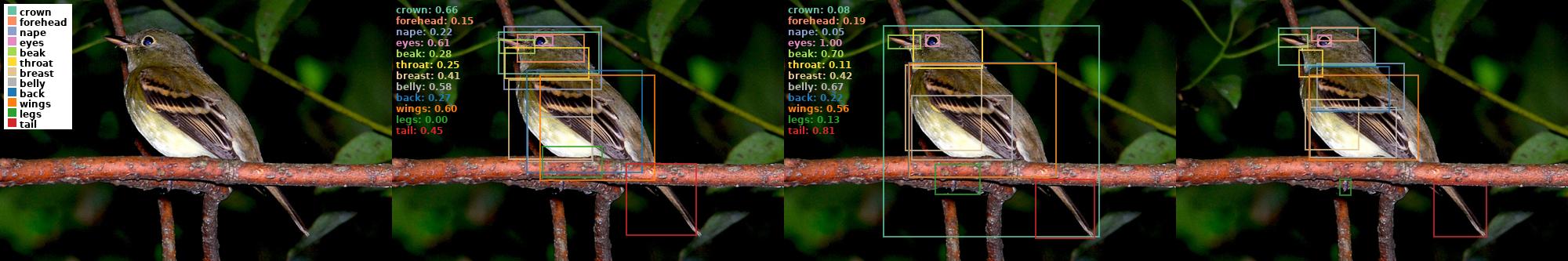}
    \end{subfigure}
    \begin{subfigure}{\textwidth}
        \centering
        \includegraphics[width=1\linewidth]{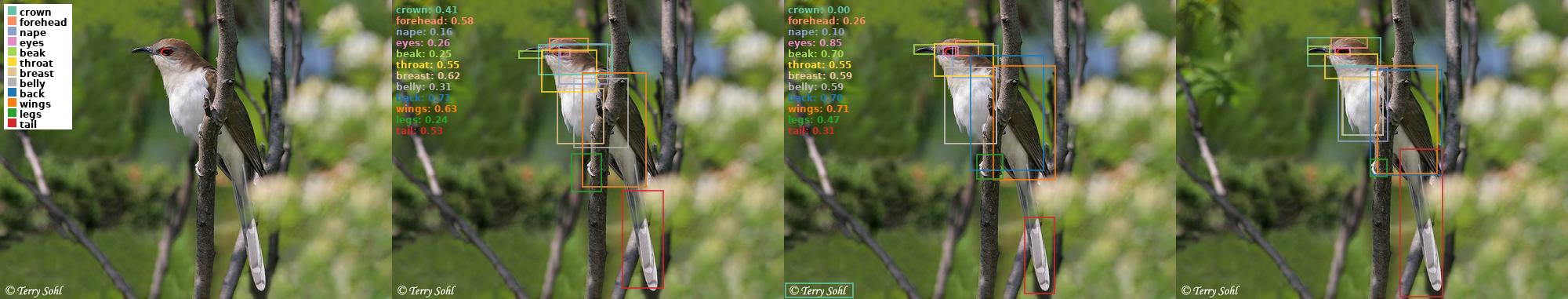}
    \end{subfigure}
    \begin{subfigure}{\textwidth}
        \centering
        \includegraphics[width=1\linewidth]{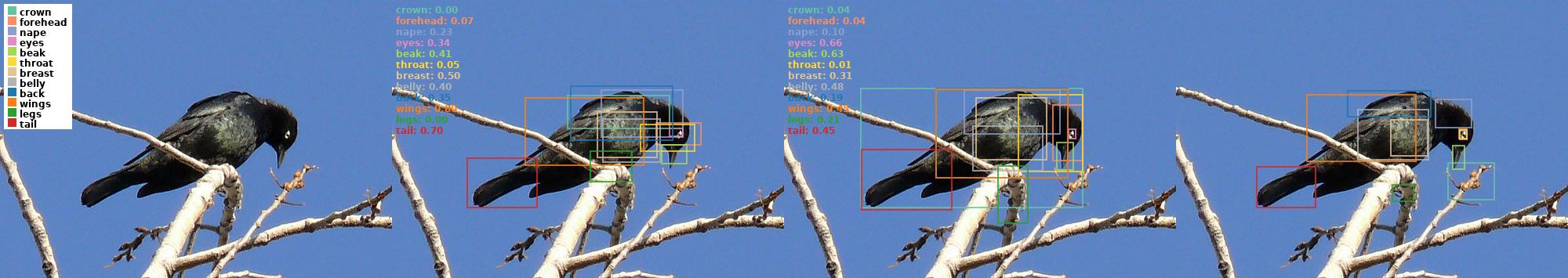}
    \end{subfigure}
    \begin{subfigure}{\textwidth}
        \centering
        \includegraphics[width=1\linewidth]{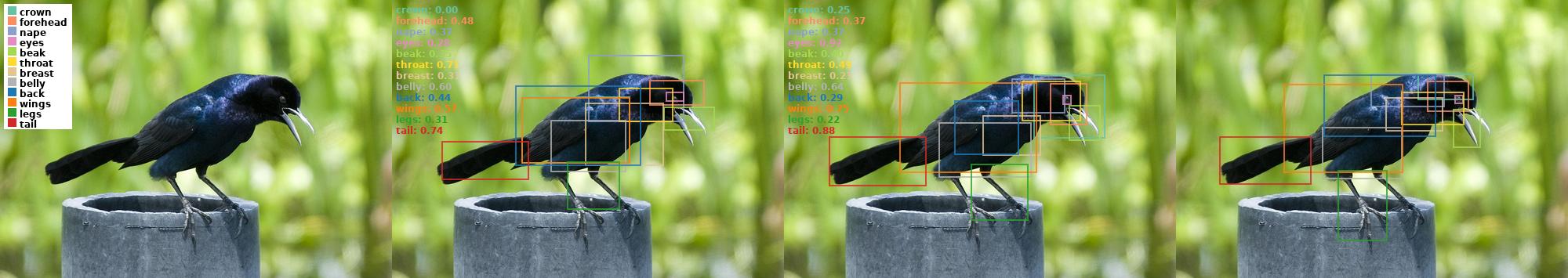}
    \end{subfigure}    
    \begin{subfigure}{\textwidth}
        \centering
        \includegraphics[width=1\linewidth]{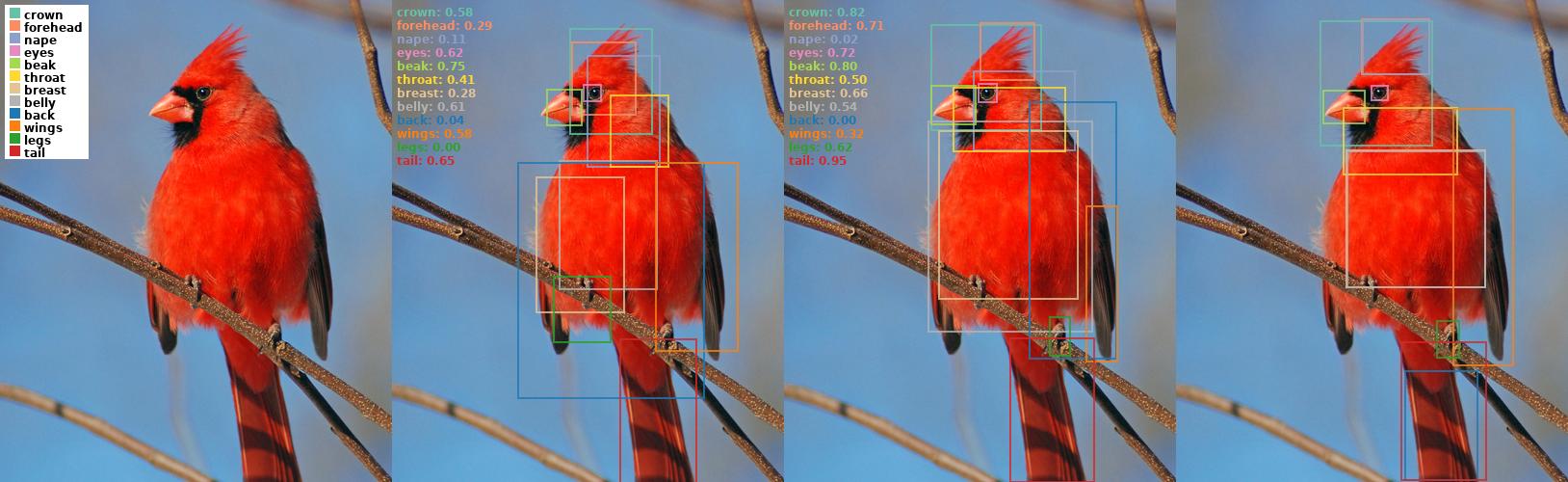}
    \end{subfigure}  
    \begin{subfigure}{\textwidth}
        \centering
        \includegraphics[width=1\linewidth]{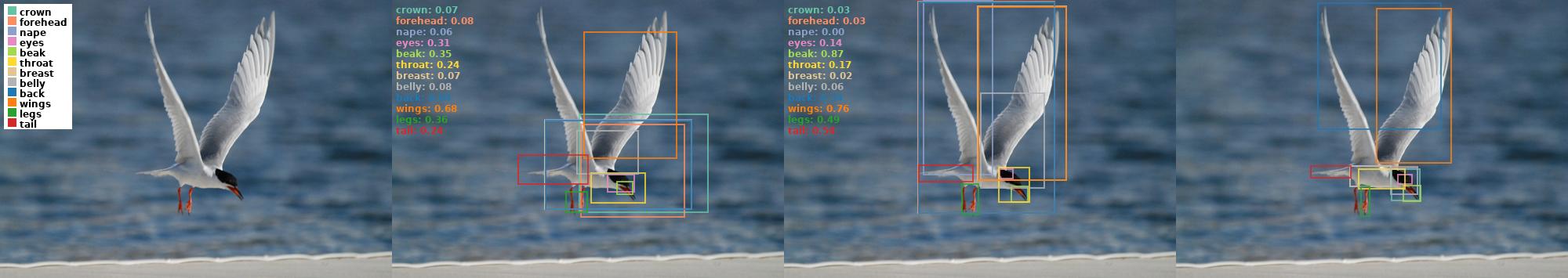}
    \end{subfigure}  
    \begin{subfigure}{\textwidth}
        \centering
        \includegraphics[width=1\linewidth]{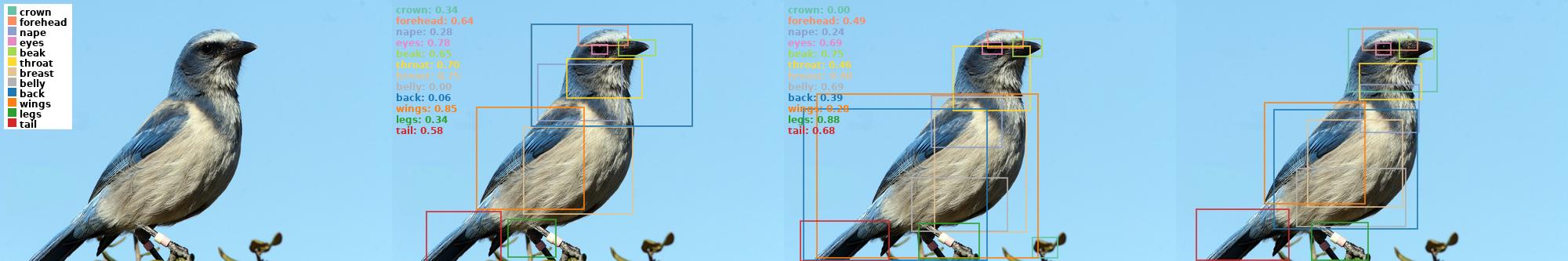}
    \end{subfigure}  
    \caption{Our predicted boxes (second column) often align closely with those of \owlvitBase (third column). However, slight shifts can lead to significant IoU discrepancies. For instance, in the first row, both \peeb and \owlvitBase accurately identify the tail. Yet, variations in focus yield a stark IoU contrast of 0.45 versus 0.81.}
    \label{fig:box_comparison}
\end{figure*}

We provide a visual comparison of the box prediction from \owlvitLarge, \owlvitBase, and \peeb in \Cref{fig:box_comparison}. We find that despite the fact that our predicted boxes have lower mean IoU compared to \owlvitLarge, they are visually similar to the boxes as \owlvitBase.

\subsection{Qualitative examples of using randomized descriptors}
\begin{figure*}
    \centering
    \tiny\makebox[0.6\textwidth][c]{Original Descriptor}%
    \tiny\makebox[0.15\textwidth][c]{Random nonsense Descriptor}%

    \begin{minipage}{0.01\textwidth}
        \rotatebox{90}{\mv}\\[70pt]
        \rotatebox{90}{\peeb}
    \end{minipage}%
    \begin{minipage}{0.95\textwidth}
        \includegraphics[width=\linewidth]{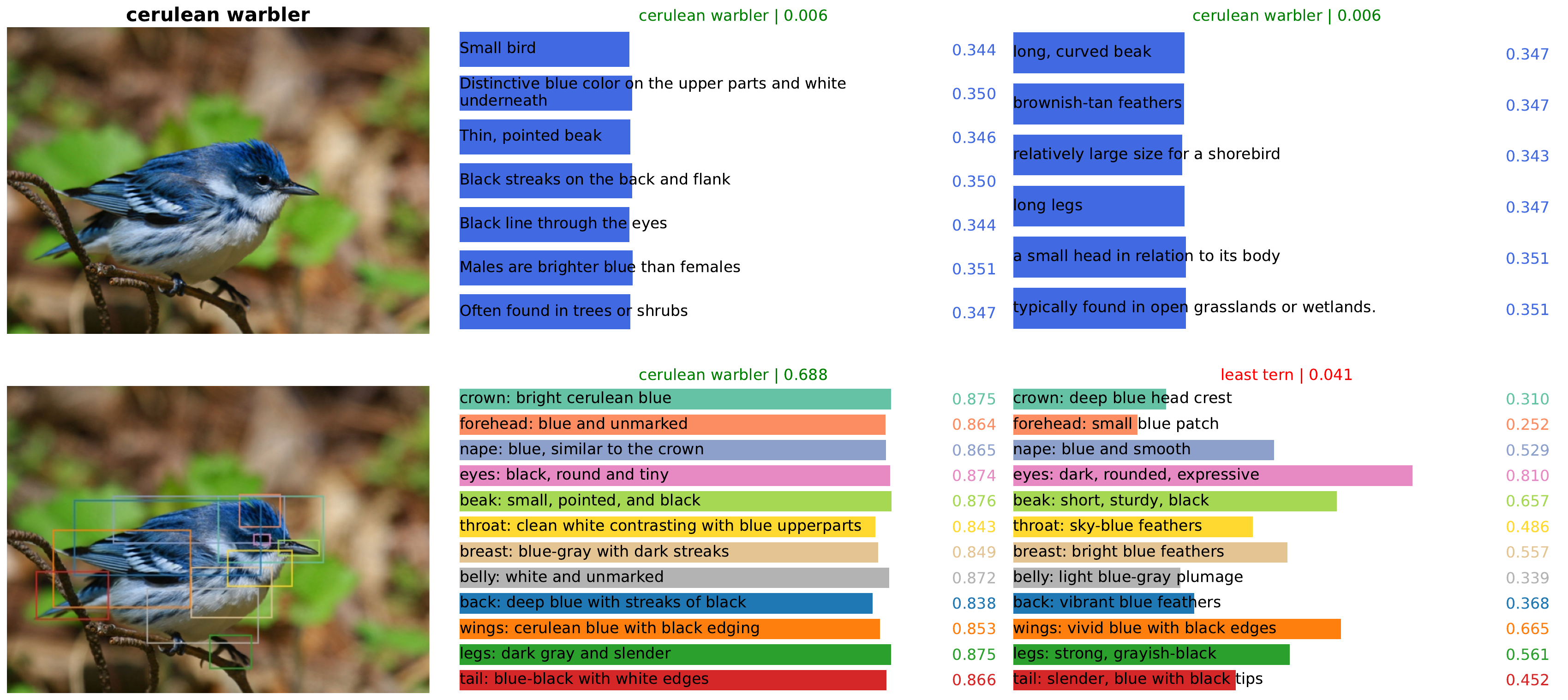}
    \end{minipage}
    \caption{Qualitative example of original descriptors vs. randomized descriptors. Upon swapping descriptors randomly, the prediction outcomes from \mv exhibit minimal variations.}
    \label{fig:randomized_description_1}
\end{figure*}

\begin{figure*}
    \centering
    \tiny\makebox[0.6\textwidth][c]{Original descriptor}%
    \tiny\makebox[0.15\textwidth][c]{Random nonsense descriptor}%

    \begin{minipage}{0.01\textwidth}
        \rotatebox{90}{\mv}\\[70pt]
        \rotatebox{90}{\peeb}
    \end{minipage}%
    \begin{minipage}{0.95\textwidth}
        \includegraphics[width=\linewidth]{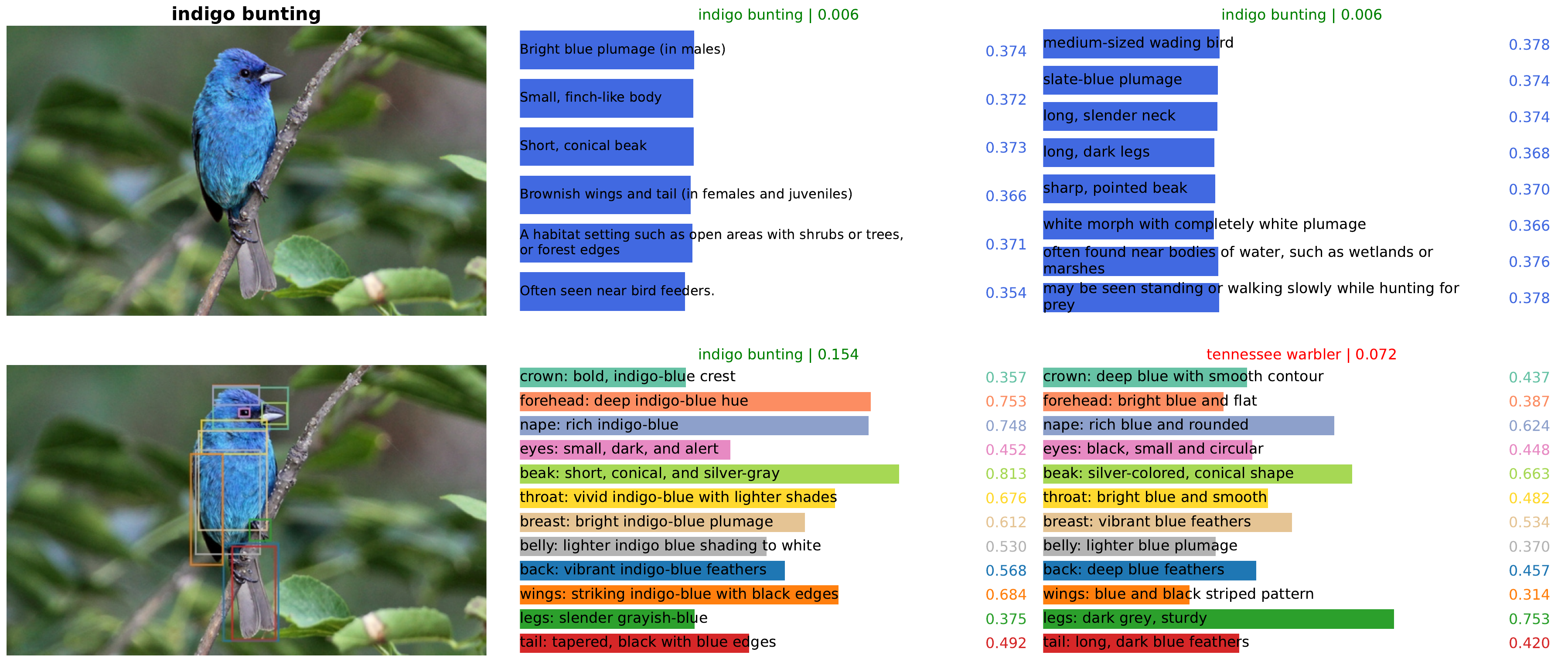}
    \end{minipage}
    \caption{Qualitative example of original descriptors vs. randomized descriptors. Since \peeb's decision is made by the descriptors, the model will try to find the descriptors that best match the image. \eg, in the random descriptors, most parts are blue.}
    \label{fig:randomized_description_2}
\end{figure*}

\begin{figure*}
    \centering
    \tiny\makebox[0.6\textwidth][c]{Original descriptor}%
    \tiny\makebox[0.15\textwidth][c]{Random nonsense descriptor}%

    \begin{minipage}{0.01\textwidth}
        \rotatebox{90}{\mv}\\[70pt]
        \rotatebox{90}{\peeb}
    \end{minipage}%
    \begin{minipage}{0.95\textwidth}
        \includegraphics[width=\linewidth]{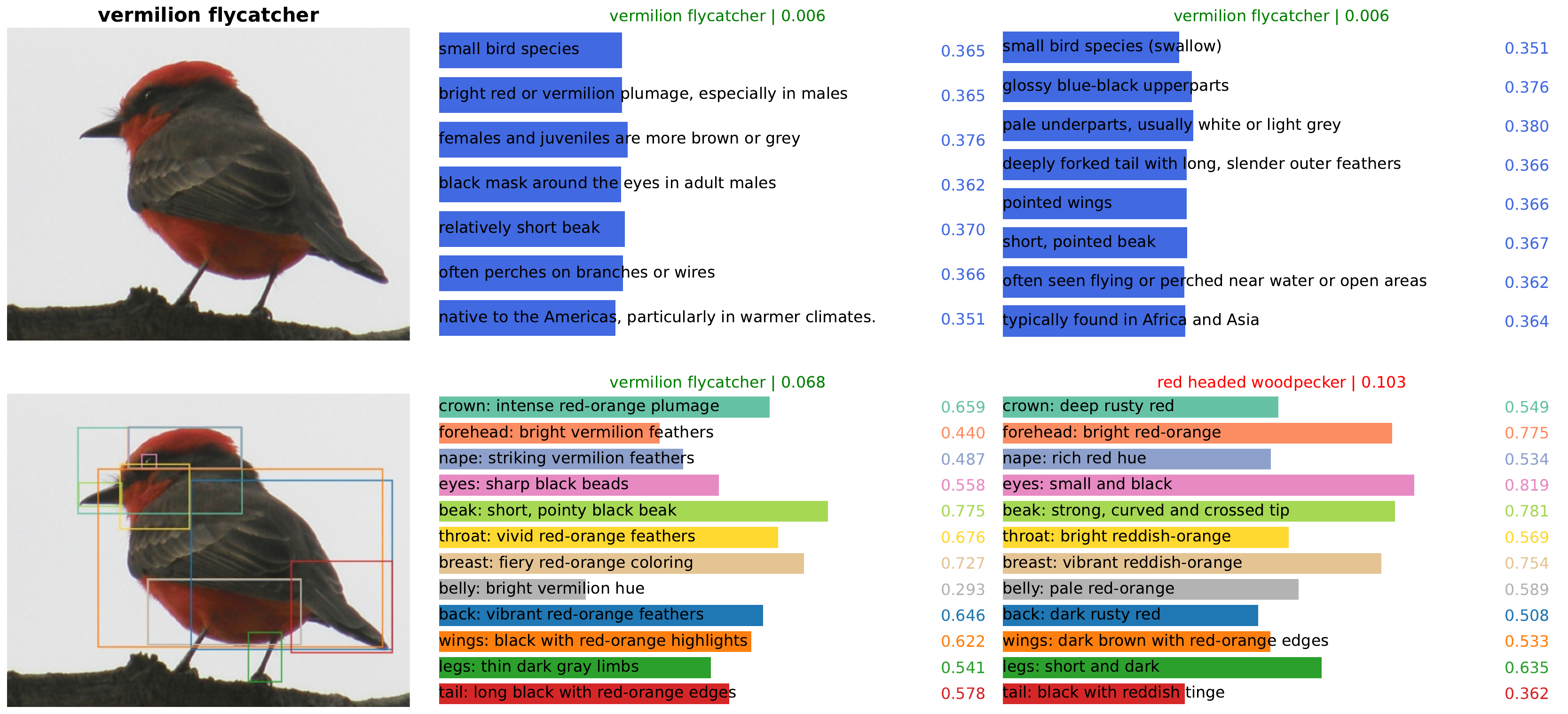}
    \end{minipage}
    
    \caption{Qualitative example of original descriptors vs. randomized descriptors. \mv maintains similar scores even for mismatched descriptors. For instance, ``bright red or vermilion plumage, especially in males'' receives a score lower than ``glossy blue-black upperparts''. Conversely, \peeb leverages the descriptors for classification, consistently relying on the descriptors that most closely align with the image.}
    \label{fig:randomized_description_3}
\end{figure*}

We visually compare \mv and \peeb based on their utilization of descriptors. (\Cref{fig:randomized_description_1,fig:randomized_description_2,fig:randomized_description_3}). 
Specifically, we randomly swap the descriptors of the classes and then use these randomized descriptors as textual inputs to the tested models to see how they perform.
We observe that the scores from \mv tend to cluster closely together. 
Surprisingly, \mv's prediction remains unchanged despite the inaccurate descriptors. 
In contrast, \peeb, when presented with randomized descriptors, attempts to identify the best match grounded on the given descriptors.

\subsection{Examples of \peeb explanations for birds}
\begin{figure*}[htb]
    \centering
    \includegraphics[width=1\linewidth]{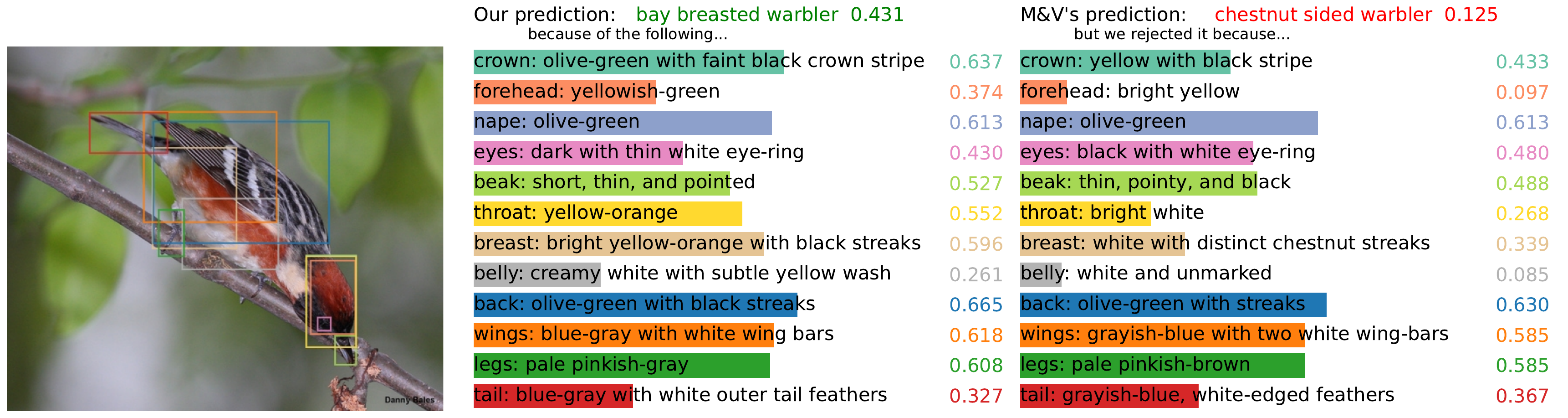}
    \caption{An example of \peeb explanation. We can see that the descriptors of these two classes are largely similar, but \peeb makes the correct prediction based on the distinctive feature of the forehead in the two classes.}
    \label{fig:exp_example1}
\end{figure*}

\begin{figure*}[htb]
    \centering
    \includegraphics[width=1\linewidth]{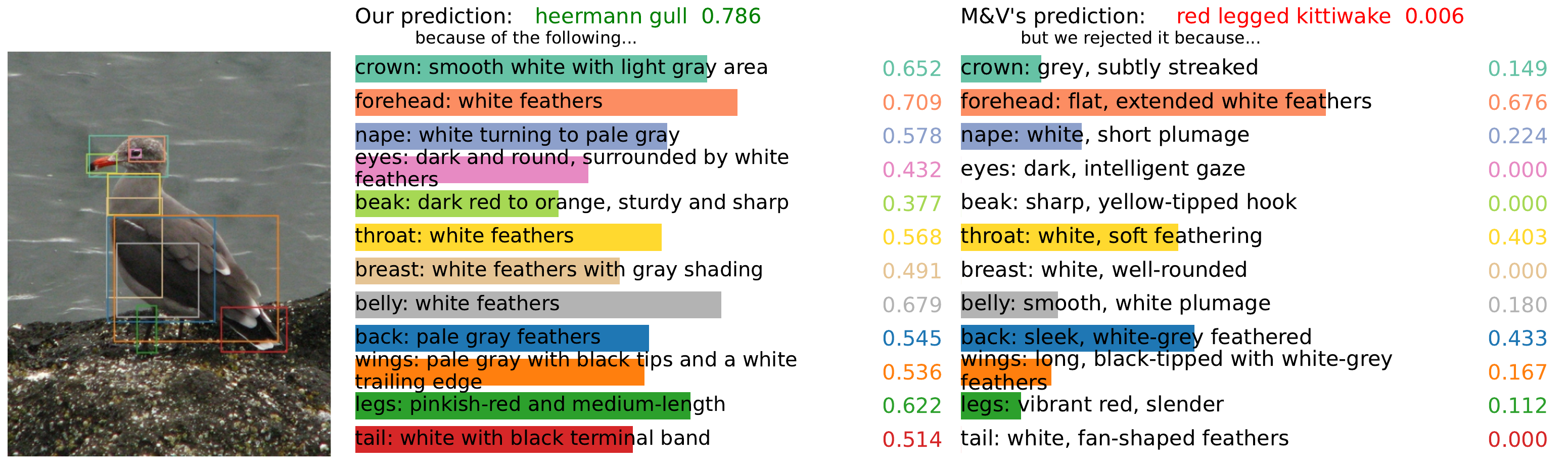}
    \caption{An example of \peeb explanation. \mv incorrectly classifies it as \textcolor{red}{\class{red-legged kittiwake}} where the \textcolor{ForestGreen}{\class{heermann gull}} does not have red legs but a red beak. 
    This example shows that CLIP is strongly biased towards some particular descriptors.}
    \label{fig:exp_example2}
\end{figure*}

\begin{figure*}[htb]
    \centering
    \includegraphics[width=1\linewidth]{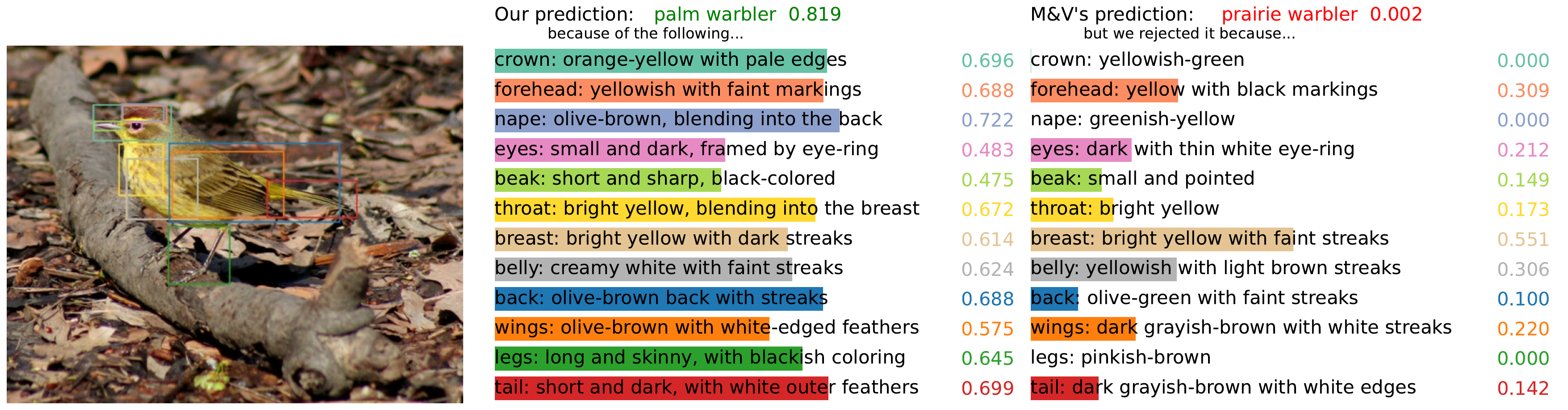}
    \caption{An example of \peeb explanation. We can see that when the descriptor does not match the image, the matching score tends to be zero, e.g., \textit{crown: yellowish-green}. 
    The clear differences in scores provide us transparency of the model's decision.}
    \label{fig:exp_example3}
\end{figure*}

\Cref{fig:exp_example1,fig:exp_example2,fig:exp_example3} are examples of how \peeb makes classification based on the descriptors and how it can reject the predictions made by \mv. 
Since we aggregate all descriptors for the final decision, even if some of them are similar in two classes, our method can still differentiate them from other descriptors. 
For instance, in \Cref{fig:exp_example1}, while other descriptors are similar, \peeb can still reject \class{chesnut-sided warbler} thanks to the distinct features of \textit{forehead}, \textit{throat} and \textit{belly}.

\subsection{Examples of \peeb explanations for dogs}
\begin{figure*}[htb]
    \centering
    \includegraphics[width=1\linewidth]{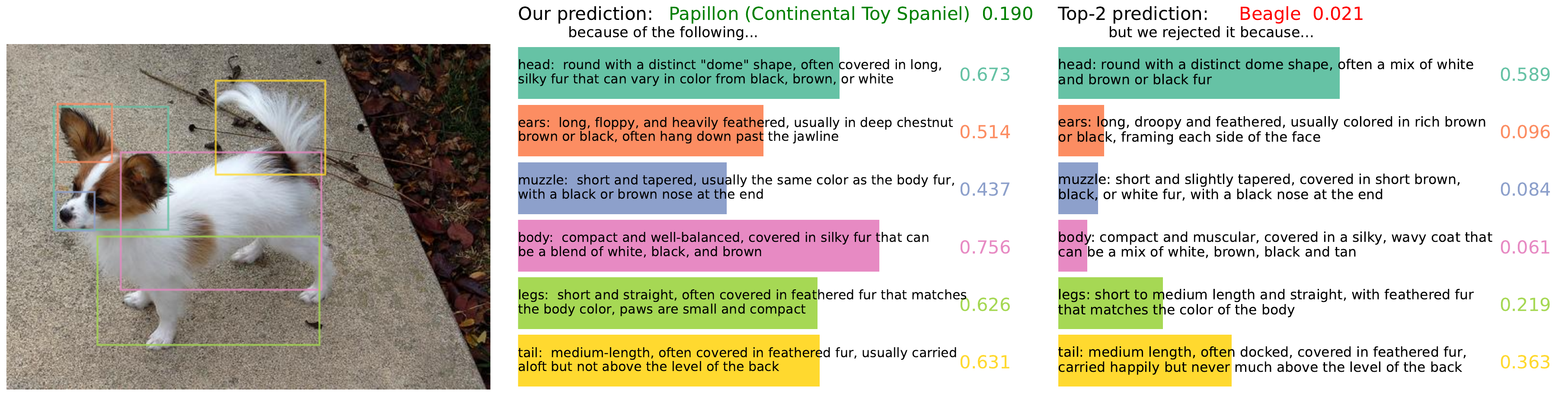}
    \caption{An example of \peeb explanation for dogs. Like birds, \peeb first identifies the predefined parts and then matches them to the descriptions.}
    \label{fig:exp_dog_example1}
\end{figure*}

\begin{figure*}[htb]
    \centering
    \includegraphics[width=1\linewidth]{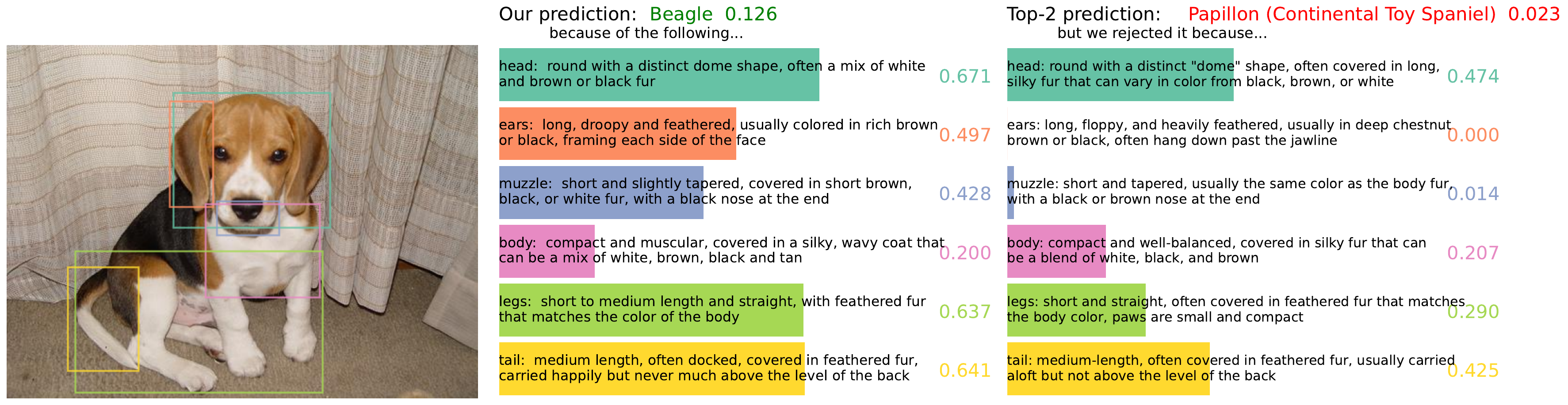}
    \caption{An example of \peeb explanation for dogs. Like birds, \peeb first identifies the predefined parts and then matches them to the descriptions.}
    \label{fig:exp_dog_example2}
\end{figure*}

\begin{figure*}[htb]
    \centering
    \includegraphics[width=1\linewidth]{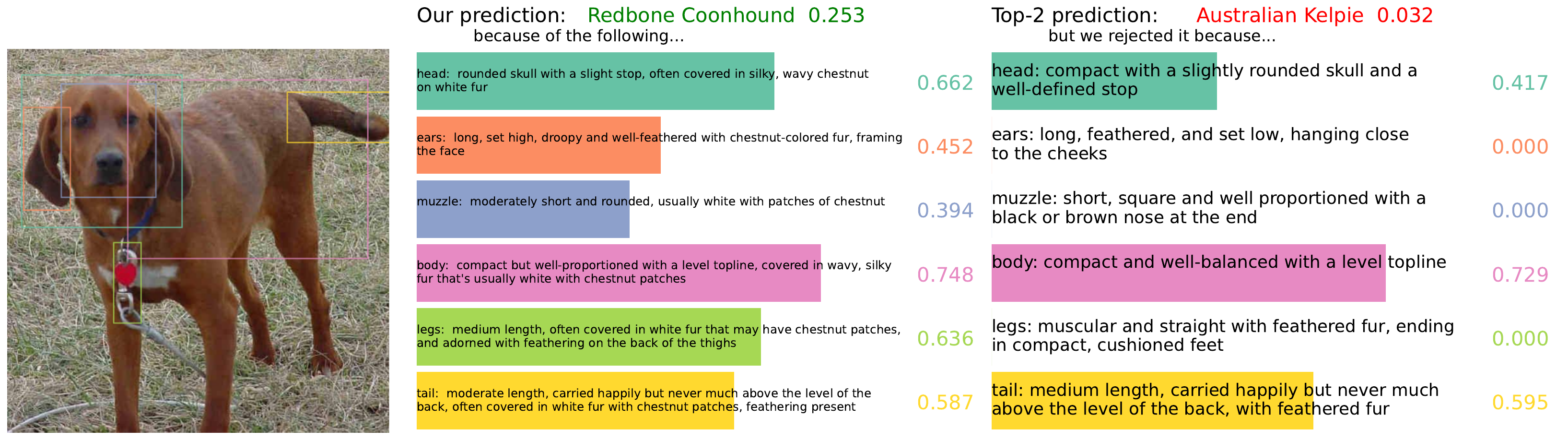}
    \caption{An example of \peeb explanation for dogs. Like birds, \peeb first identifies the predefined parts and then matches them to the descriptions.}
    \label{fig:exp_dog_example5}
\end{figure*}

\Cref{fig:exp_dog_example1,fig:exp_dog_example2,fig:exp_dog_example5} are examples of how \peeb makes classification based on the descriptors in Stanford Dogs dataset. We demonstrate that our model works well on dogs, which indicates that our proposed method is transferable to other domains while maintaining high-quality explainability as in birds.

\end{document}